\documentclass{article}

\PassOptionsToPackage{table,xcdraw}{xcolor}
\usepackage{xcolor}
\def\isaccepted{} 
\usepackage[final]{corl_2026} 
\hypersetup{pdftitle={ActionCodec: What Makes for Good Action Tokenizers}, pdfauthor={Zibin Dong, Yicheng Liu, Shiduo Zhang, Baijun Ye, Yifu Yuan, Fei Ni, Jingjing Gong, Xipeng Qiu, Hang Zhao, Yinchuan Li, Jianye Hao}}

\makeatletter
\def\@maketitle{%
  \vbox{%
    \hsize\textwidth
    \linewidth\hsize
    \vskip 0.1in
    \centering
    {\LARGE\bf \@title\par}
    \if@conferencefinal
      \def\And{\end{tabular}\penalty0\hspace{3pt}\begin{tabular}[t]{c}\bf\rule{\z@}{11\p@}\ignorespaces}%
      \def\AND{\end{tabular}\hfil\linebreak[4]\hfil\begin{tabular}[t]{c}\bf\rule{\z@}{11\p@}\ignorespaces}%
      \begin{tabular}[t]{c}\bf\rule{\z@}{24\p@}\@author\end{tabular}%
    \else
      \if@preprinttype
        \def\And{\end{tabular}\penalty0\hspace{3pt}\begin{tabular}[t]{c}\bf\rule{\z@}{11\p@}\ignorespaces}%
        \def\AND{\end{tabular}\hfil\linebreak[4]\hfil\begin{tabular}[t]{c}\bf\rule{\z@}{11\p@}\ignorespaces}%
        \begin{tabular}[t]{c}\bf\rule{\z@}{24\p@}\@author\end{tabular}%
      \else
        \begin{tabular}[t]{c}\bf\rule{\z@}{24\p@}
          Anonymous Author(s) \\
          Affiliation \\
          Address \\
          \texttt{email} \\
        \end{tabular}%
      \fi
    \fi
    \vskip 0.18in \@minus 0.08in
  }%
}
\makeatother

\usepackage{microtype}
\usepackage{graphicx}
\usepackage{subcaption}
\usepackage{amsmath}
\usepackage{bm}
\usepackage{amssymb}
\usepackage{mathtools}
\usepackage{amsthm}
\usepackage{booktabs}
\usepackage{multirow}
\usepackage{array}
\usepackage{colortbl, makecell}
\usepackage{tcolorbox}
\usepackage{wrapfig}
\usepackage{enumitem}
\usepackage{pifont}
\usepackage[capitalize,noabbrev]{cleveref}

\theoremstyle{plain}

\theoremstyle{definition}

\theoremstyle{remark}

\title{ActionCodec: What Makes for Good Action Tokenizers}

\author{
Zibin Dong$^{1,2,*,\ddagger}$ \And
Yicheng Liu$^{1,*}$ \And
Shiduo Zhang$^{4,5}$ \And
Baijun Ye$^{1}$ \And
Yifu Yuan$^{3}$ \AND
Fei Ni$^{3}$ \And
Jingjing Gong$^{5}$ \And
Xipeng Qiu$^{4,5}$ \And
Hang Zhao$^{1,\dagger}$ \And
Yinchuan Li$^{2}$ \And
Jianye Hao$^{3,\dagger}$ \AND
{\normalfont\footnotesize $^1$Tsinghua University \enspace $^2$Knowin AI \enspace $^3$Tianjin University}\\
{\normalfont\footnotesize $^4$Fudan University \enspace $^5$Shanghai Innovation Institute}\\
{\normalfont\footnotesize $^*$Equal contribution \enspace $^\ddagger$Work completed during an internship at Knowin AI}\\
{\normalfont\footnotesize $^\dagger$Corresponding authors: \texttt{hangzhao@mail.tsinghua.edu.cn}, \texttt{jianye.hao@tju.edu.cn}}
}

\begin{document}
\setlength{\tabcolsep}{1pt} 
\maketitle
\setlength{\tabcolsep}{6pt} 

\vspace{-10pt}
\begin{abstract}
Vision-Language-Action (VLA) models leveraging the native autoregressive paradigm of Vision-Language Models (VLMs) have demonstrated superior instruction-following and training efficiency. Central to this paradigm is action tokenization, yet its design has primarily focused on reconstruction fidelity, failing to address its direct impact on VLA optimization. Consequently, the fundamental question of \textit{what makes for good action tokenizers} remains unanswered. In this paper, we bridge this gap by establishing design principles specifically from the perspective of VLA optimization. We identify a set of best practices based on information-theoretic insights, including maximized temporal token overlap, minimized vocabulary redundancy, improved multimodal alignment, and reduced architectural token coupling. Guided by these principles, we introduce \textbf{ActionCodec}, a high-performance action tokenizer that significantly enhances both training efficiency and VLA performance across diverse simulation and real-world benchmarks. Notably, on LIBERO, a SmolVLM2-2.2B fine-tuned with ActionCodec achieves a 95.5\% success rate without any VLA pre-training. With advanced architectural enhancements, this reaches 97.4\%, representing a new SOTA for VLA models without VLA pre-training. We believe that our established design principles, alongside the released model, will provide a clear roadmap for the community to develop more effective action tokenizers.
\end{abstract}


\begin{wrapfigure}{R}{0.50\textwidth}
    \vspace{-15pt}
    \centering
    \includegraphics[width=\linewidth]{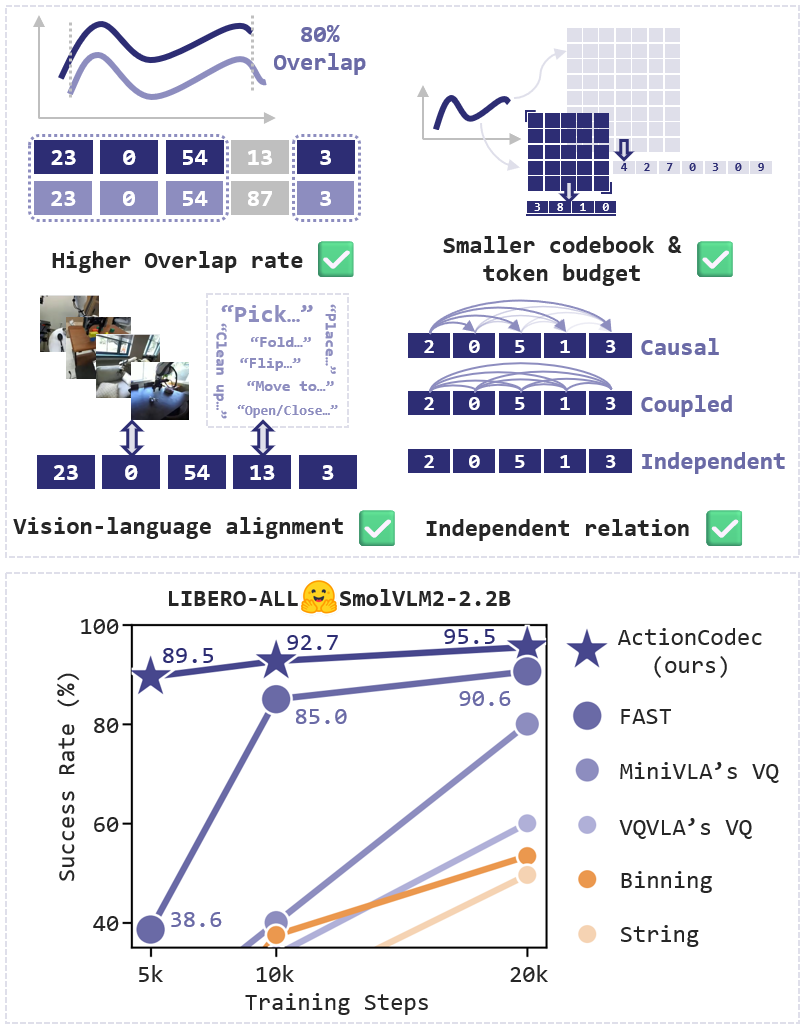}
    \caption{\textbf{ActionCodec} provides a comprehensive analysis of VQ action tokenizers that directly impact VLA training and summarizes the best practices. When utilized for the fine-tuning of SmolVLM2-2.2B without additional architectural designs, ActionCodec achieves performance on LIBERO benchmark that far exceeds other tokenizers, particularly in terms of training efficiency.}
    \label{fig:teaser}
    \vspace{-20pt}
\end{wrapfigure}

\section{Introduction}
\vspace{-5pt}

Vision-Language-Action (VLA) models, which fine-tune Vision-Language Models (VLMs) on large-scale robotic interaction datasets, have emerged as a cornerstone for general-purpose physical intelligence \citep{driess2023palm, zitkovich2023rt2, bai2024survey}. While derived from VLMs, contemporary VLA architectures often diverge from the standard autoregressive paradigm by employing specialized action heads \citep{kim2025openvlaoft, bu2025univla} or external diffusion-based experts \citep{black2024pi0, intelligence2025pi05}. Recent evidence suggests that adhering to the native VLM generative framework, representing actions as discrete tokens, is crucial for preserving pre-trained general knowledge and fostering superior instruction-following capabilities \citep{driess2025KI, intelligence2025pi06, hancock2025vlm2vla, zhang2025vlabench, dong2025cocos}. This trend is exemplified by state-of-the-art (SOTA) models like $\pi_{0.5}$ \citep{intelligence2025pi05}, where discretized action tokens are central to the fine-tuning task. Consequently, the design of an effective action tokenization scheme has become a critical bottleneck in VLA research.

Existing action tokenizers generally fall into three categories: (1) {Heuristic methods}, which directly discretize signals into bins or raw strings \citep{kim2024openvla, goyal2025vla0}; (2) {Semi-data-driven methods}, which apply Byte-Pair Encoding (BPE) to frequency-domain signals \citep{pertsch2025fast}; and (3) {Data-driven methods}, primarily based on Vector Quantization (VQ) to learn latent discrete representations \citep{wang25vqvla}. Although VQ-based approaches are theoretically superior due to their minimal reliance on heuristic priors, they have historically lagged behind simpler schemes in downstream robotic tasks. We attribute this discrepancy to a fundamental research gap: existing literature evaluates VQ tokenizers through the lens of generative fidelity, neglecting the complex training dynamics of the VLA backbone. 

In this paper, we address the foundational question: \textit{What makes for good action tokenizers} for VLA optimization? Moving beyond standard generative metrics, we analyze VLA performance through the prisms of information theory and representation learning. We identify four key desiderata for high-performance action tokens: (i) maximized temporal overlap between adjacent chunks, (ii) minimized vocabulary redundancy, (iii) improved alignment between tokens and multimodal contexts, and (iv) reduced architectural token coupling.

Guided by these principles, we introduce \textbf{ActionCodec}, a robust action tokenizer that integrates the empirically validated design choices. Moreover, ActionCodec leverages Residual Vector Quantization (RVQ) \citep{lee2022rvq} post-training to refine reconstruction fidelity and incorporates embodiment-specific soft prompts to facilitate knowledge transfer across diverse robotic platforms. Our empirical results demonstrate that VLAs equipped with ActionCodec, without any additional architectural modifications, achieve improvements in training efficiency, success rates, and anti-overfitting resilience across diverse benchmarks. ActionCodec achieves SOTA performance in both simulated and real-world environments, providing a systematic roadmap for the future of VQ-based action representation. Our contributions are summarized as follows:

\begin{itemize}[leftmargin=*, itemsep=0pt, topsep=2pt]
    \item We establish an information-theoretic framework to motivate practical design desiderata for action tokenization from the perspective of VLA training dynamics, providing a systematic roadmap for future research in general-purpose physical intelligence.
    \item We introduce \textbf{ActionCodec}, a high-performance action tokenizer that instantiates these validated principles. ActionCodec greatly enhances the VLA training efficiency and mitigates overfitting.
    \item We provide a comprehensive suite of ActionCodec-powered VLA models across diverse architectural paradigms. Our approach achieves SOTA performance on multiple simulated and real-world benchmarks, showing remarkable efficacy even initialized from VLMs without VLA pre-training.
\end{itemize}

\begin{figure}[t]
    \centering
    \includegraphics[width=0.95\linewidth]{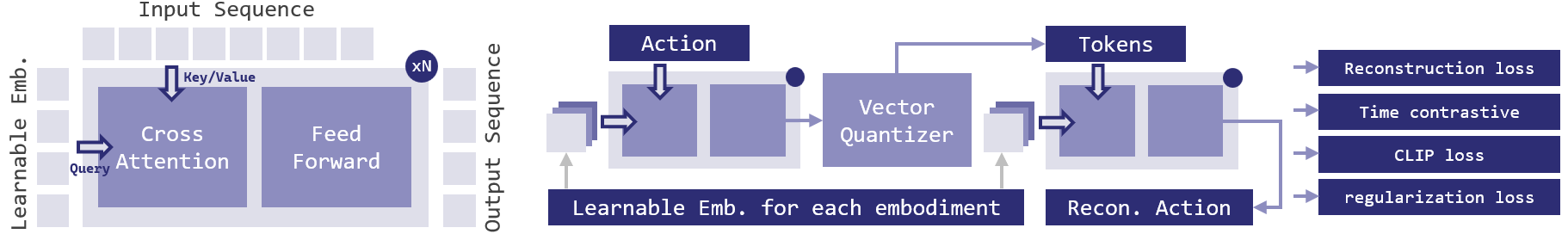}
    \caption{{\small\textbf{Neural Network Architecture of ActionCodec.} We employ a Perceiver-like transformer architecture due to its inherent flexibility, which facilitates the modeling of diverse token relations and supports the encoding of variable-length action sequences.}}
    \label{fig:arch}
    \vspace{-10pt}
\end{figure}

\vspace{-5pt}
\section{Preliminaries}
We consider learning a policy $\pi_\theta(A | V, L)$ that maps visual observations $V$ and language instructions $L$ to an action sequence $A \in \mathbb{R}^{T \times D}$. To leverage the autoregressive paradigm of pre-trained VLMs, $A$ is discretized into a proxy sequence of tokens $C = \left[c_1, c_2, \dots, c_n\right]$, where each $c_k$ belongs to a vocabulary $\mathcal{K} = \{1, \dots, S\}$. This process is facilitated by a Vector Quantized (VQ) \citep{oord2017vqvae} tokenizer comprising an encoder $\mathcal{F}$ and a decoder $\mathcal{G}$. Specifically, $\mathcal{F}$ maps $A$ to a continuous latent representation $Z = [{z}_1, \dots, {z}_n] \in \mathbb{R}^{n \times d}$, which is then quantized against a learnable codebook $\mathcal{B} = \{{e}_j\}_{j=1}^S \subset \mathbb{R}^d$ such that $c_k = \arg \min_{j \in \{1, \dots, S\}} \|{z}_k - {e}_j\|_2$. The reconstructed action sequence is denoted as $\hat{{A}} = \mathcal{G}({e}_{c_1}, \dots, {e}_{c_n})$. The tokenizer is optimized via 

\vspace{-5pt}
\begin{equation}
    \mathcal{L}_{VQ} = \|{A} - \hat{{A}}\|_2^2 + \|\text{sg}[{Z}] - {e}_{{c}}\|_2^2 + \|{Z} - \text{sg}[{e}_{{c}}]\|_2^2,
\end{equation}
where $\text{sg}[\cdot]$ is the stop-gradient operator.

\section{A VLA Perspective on Action Tokenization}\label{sec:before_method}

We formalize action tokenization as the construction of a practical proxy target for VLA optimization. Since the tokenizer dictates the target distribution, VLA training involves both model fitting and the implicit optimization of the supervision signal. The expected negative log-likelihood (NLL) loss is:
\begin{equation}
    \mathbb{E}\left[\mathcal{L}_{\text{NLL}}\right] = D_{KL}(P_{\text{data}} \parallel P_\theta) + H(C|V,L),
\end{equation}
where $H(C|V,L)$ quantifies \textit{supervisory ambiguity}. This exact identity separates target entropy from model-fitting error. The following exact entropy decomposition provides an analytical lens, rather than a guarantee that optimizing each term improves policy performance:
\begin{equation}\label{eq:entropy_decomposition}
    H(C | V, L) = \underbrace{H(C | A)}_{\text{\tiny Deterministic Term}} + \underbrace{I(C; A)}_{\text{\tiny Capacity}} - \underbrace{I(C; V, L)}_{\text{\tiny Perceptual Alignment}}
\end{equation}
Guided by \Cref{eq:entropy_decomposition}, we study three aspects empirically. \textbf{(a) Overlap Rate.} Deterministic tokenization gives $H(C|A)=0$. Local instability instead concerns perturbation-induced entropy $H_{\epsilon}(\mathcal{T}(A+\epsilon)|A)$, where $\mathcal{T}$ includes encoding and quantization. We use the \textit{overlap rate} (OR) between adjacent chunks as a stability proxy, not a numerical entropy estimate (\Cref{appendix:artifact_entropy,appendix:or_controls}). \textbf{(b) Capacity and Vocabulary Size.} For deterministic tokens, $I(C;A)=H(C)\le\sum_{k=1}^n H(c_k)\le n\log_2 S$. This exact bound depends on token budget $n$ and vocabulary size $S$. Our practical objective is to reduce $H(C)$ subject to adequate reconstruction fidelity (\Cref{appendix:capacity_scope}); smaller budgets are not universally better. \textbf{(c) Perceptual Alignment vs. Residual Grammar.} To understand the information utilized for prediction, we decouple the total information gain for a token $c_k$ using the chain rule:
\begin{equation}
    \underbrace{I(c_k; V, L, c_{<k})}_{\text{\tiny Total Information}}
    = \underbrace{I(c_k; V, L)}_{\text{\tiny Visual-Language Alignment}} 
    + \underbrace{I(c_k; c_{<k} | V, L)}_{\text{\tiny Residual Grammar}}
\end{equation}
\textit{Visual-Language alignment} measures the direct correlation between tokens and multimodal inputs, while \textit{residual grammar} quantifies the dependency on autoregressive history given the context. These complementary pathways motivate empirical studies of context alignment and prefix-error robustness; neither mutual information is directly estimated (\Cref{appendix:alignment_scope}). 


Guided by these principles, the subsequent sections show the derivation of best practices for VQ-based tokenization, starting from a vanilla VQ-VAE architecture (\Cref{fig:arch}) trained on the LIBERO \citep{liu2024libero} dataset as an empirical foundation. To validate the efficacy of these design choices, we move beyond intrinsic reconstruction metrics and directly fine-tune SmolVLM2-256M \citep{marafioti2025smolvlm} as a VLA model. The quality of the tokenization scheme is thus measured by the resulting policy's performance on the LIBERO-Goal benchmark (refer to \Cref{appendix:validation_exp} for comprehensive experimental settings).

\subsection{Impacts of Overlap Rate}\label{sec:impacts_or}

\begin{wrapfigure}{r}{0.45\textwidth}
    \vspace{-10pt}
    \centering
    \includegraphics[width=\linewidth]{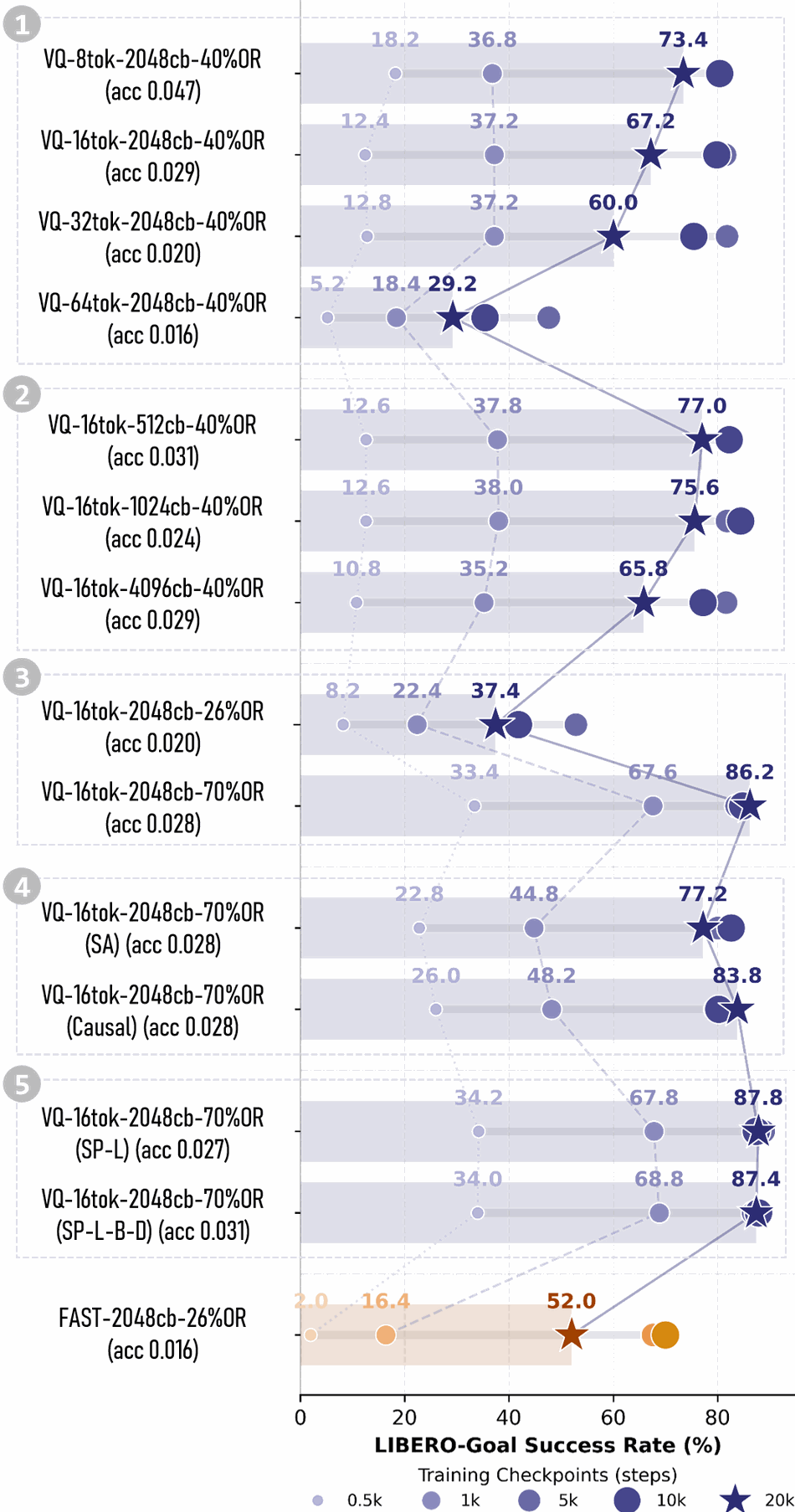}
    \vspace{-15pt}
    \caption{{\small\textbf{LIBERO-Goal results for different design choices.} All VLA models are based on SmolVLM2-256M, following vocabulary expansion and full-parameter fine-tuning without additional architectural modifications. The suffix notations are defined as follows: \texttt{acc} (L1 error), \texttt{tok} (token budget), \texttt{cb} (vocabulary size), \texttt{OR} (overlap rate), \texttt{SA} (self-attention), \texttt{Causal} (SA w/ causal mask), and \texttt{SP} (training w/ soft-prompt).}}
    \label{fig:toy_example_ablation}
    \vspace{-20pt}
\end{wrapfigure}

To investigate the effects of the overlap rate (OR) on VLA optimization, we synthesize three tokenizer variants with controlled OR levels ($26\%$, $40\%$, and $70\%$) by augmenting the VQ-VAE objective with a contrastive loss to modulate the topological stability of the token space. The performance on LIBERO-Goal, summarized in \Cref{fig:toy_example_ablation} \textcolor{gray}{(Block 1 \& 3)}, demonstrates that a higher OR consistently

enhances training efficiency, asymptotic convergence, and robustness. Notably, the $70\%$ OR tokenizer enables the VLA to achieve a $33.4\%$ success rate within a mere 500 training steps, significantly outperforming the naive VQ-VAE ($8.2\%$) and the FAST baseline ($2\%$). To further explore the underlying mechanism, we select data from four tasks and visualize the last hidden states of the VLA's action $\texttt{[BOS]}$ token using t-SNE, as shown in the first three rows of \Cref{fig:tsne_visualization}. The visualizations reveal that a high OR enables the VLA model to form distinct clusters in the latent space earlier in the training process. Furthermore, the cluster distributions remain more stable over training, consistent with reduced late-stage rollout degradation (\Cref{appendix:optimization_metrics}). {Consequently, we establish maximizing the overlap rate as a primary design requirement for the action tokenizer.}

\subsection{Impacts of Vocabulary Size and Token Budget}
\vspace{-5pt}

As established in our theoretical framework, the capacity of the action proxy is governed by the entropy upper bound $n \log_2 S$. We evaluate various configurations spanning $n \in \{8, 16, 32, 64\}$ and $S \in \{512, 1024, 2048, 4096\}$ with OR held at 40\% by adjusting the contrastive weight to study their influence on optimization. Our results in \Cref{fig:toy_example_ablation} \textcolor{gray}{(Blocks 1 \& 2)} indicate that these parameters primarily dictate the VLA's resilience to overfitting, with the token budget $n$ exerting a substantially more dominant influence than the vocabulary size $S$. A comparison across Rows 1, 4, and 5 of \Cref{fig:tsne_visualization} further elucidates this phenomenon: while an increased token budget $n$ preserves sufficient task discriminability, it leads to excessive dispersion in the latent clusters. This expanded representational capacity makes the model increasingly susceptible to capturing high-frequency noise and spurious correlations. Conversely, aggressive reduction of $n$ and $S$ eventually incurs a non-trivial reconstruction penalty, as the tokenizer fails to capture the necessary physical nuances of the action space. By navigating this trade-off between fidelity and the learning manifold complexity, we identify $S=2048$ and $n=16$ as a practical configuration, rather than a unique optimum.

\vspace{-5pt}
\subsection{Impacts of Vision-Language Alignment and Residual Grammar}\label{sec:vl_align_and_grammar}
\vspace{-5pt}

\begin{figure}[t]
    \centering
    \vspace{-10pt}
    \begin{minipage}[t]{0.38\textwidth}
        \centering
        \includegraphics[width=\linewidth]{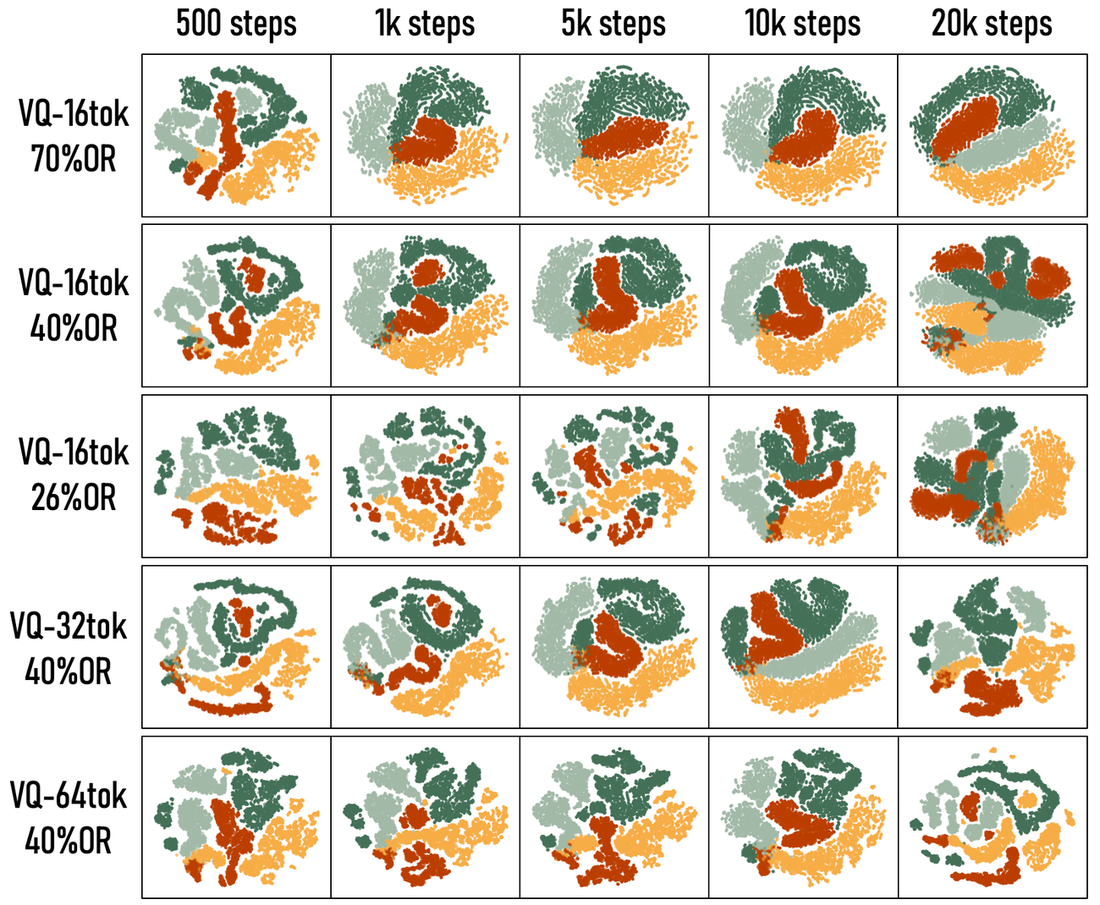}
        \caption{\small t-SNE visualization of the VLA's last hidden states for the action $\texttt{[BOS]}$ token on four LIBERO-Goal tasks.}
        \label{fig:tsne_visualization}
    \end{minipage}\hfill
    \begin{minipage}[t]{0.58\textwidth}
        \centering
        \includegraphics[width=\linewidth]{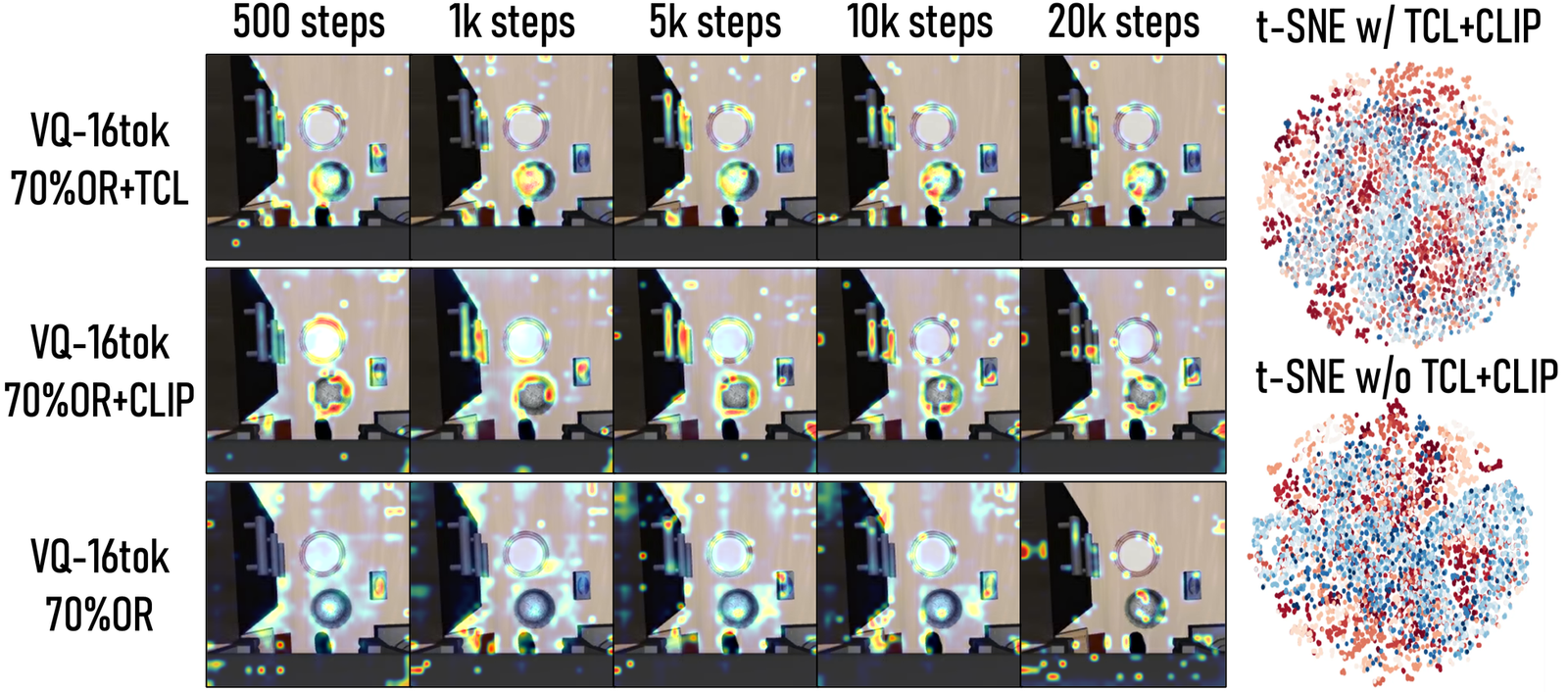}
        \caption{\small (Left) VLA attention maps for CLIP vs. TCL-trained tokenizers. (Right) t-SNE visualization of the structured latent space across 40 LIBERO tasks.}
        \label{fig:vl_align}
    \end{minipage}
    \vspace{-10pt}
\end{figure}

Within the autoregressive framework, we decompose the token's perceptual alignment into two competing pathways: \textit{visual-language alignment} $I(c_k; V, L)$, representing environmental grounding, and \textit{residual grammar} $I(c_k; c_{<k} | V, L)$, quantifying autoregressive dependencies. To optimize the former, we integrate Time Contrastive Learning (TCL) and CLIP-based objectives into the training:
\begin{align}
    \mathcal{L}_{\text{TCL}} &= -\sum_{i}\log \frac{e^{\mathcal S(Z^i, Z^{i+})}}{e^{\mathcal S(Z^i, Z^{i+})} + e^{\mathcal S(Z^i, Z^{i-})}}, \\
    \mathcal{L}_{\text{CLIP}} &= -\sum_i \sum_j \log \frac{1}{1 + e^{\left(l_{ij}(-t Z^i y^j + b)\right)}},
\end{align}
where $Z^i, Z^{i+}$, and $Z^{i-}$ denote the continuous embeddings of the action chunk, its adjacent neighbor, and other action chunks within the batch, respectively; $S$ denotes cosine similarity; $y^j$ is the language embedding, $l_{ij}$ is an indicator variable that equals 1 if $i$ and $j$ are paired and -1 otherwise, and $t$ and $b$ are learnable parameters. Additionally, an $L_1$ penalty regulation loss is applied to ensure training stability. Our findings are two-fold: \textbf{Differential effects of TCL and CLIP.} Visualizations of the VLA attention maps for the action $\texttt{[BOS]}$ token (\Cref{fig:vl_align}, left) reveal that CLIP-trained tokenizers cause the VLA to focus on instruction-relevant objects, whereas TCL-trained tokenizers bias attention toward dataset-specific demonstration patterns (e.g., specific grasp points). (b) \textbf{Improved Overlap Rate and Structure.} Both objectives enhance the overlap rate and stability. As shown in the t-SNE visualizations of all 40 LIBERO tasks (\Cref{fig:vl_align}, right), TCL and CLIP training produces more structured latent clusters than simple action-perturbation methods. Thus, we adopt TCL and CLIP as the primary practices for improving OR.



\begin{wrapfigure}{r}{0.40\textwidth}
    \vspace{-15pt}
    \centering
    \includegraphics[width=\linewidth]{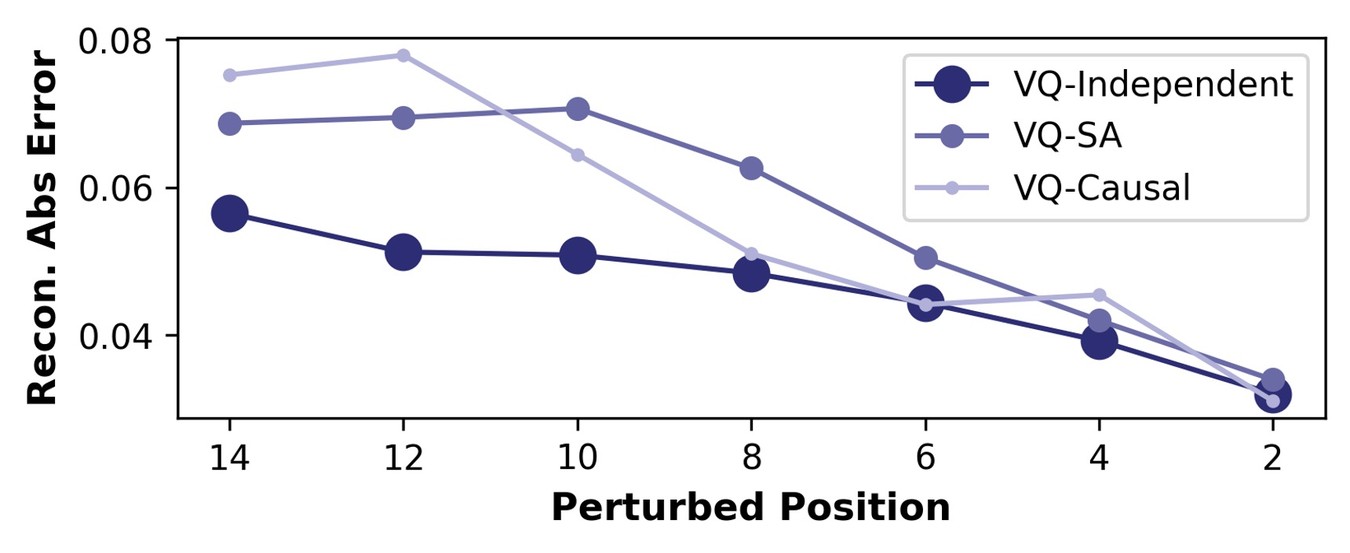}
    \vspace{-20pt}
    \caption{{\small \textbf{Error propagation under token perturbation.} We measure the L1 reconstruction error when noise is injected at specific positions during generating. VQ-Independent maintains a stable, lower error compared to SA and Causal architectures. This supports reducing architectural token coupling for prefix-error robustness.}}
    \label{fig:grammar}
    \vspace{-12pt}
\end{wrapfigure}

Regarding \textbf{Residual Grammar}, we compare Perceiver-based (cross-attention only), Self-Attention (SA), and Causal-SA tokenizers. Here, \textit{independent} denotes the absence of query self-attention, not statistical independence: all queries still attend to the same action. In the prefix-error experiment (\Cref{fig:grammar}), one predicted token is shuffled before autoregressive generation continues. The cross-attention-only variant is more robust to this perturbation and yields higher success rates and faster early convergence (\Cref{fig:toy_example_ablation}, \textcolor{gray}{Block 4}). These results support reducing coupling in the architectures studied, rather than a universal independence requirement.

\vspace{-10pt}
\section{ActionCodec}\label{sec:actioncodec}
\vspace{-5pt}

Synthesizing the aforementioned design principles, we introduce \textbf{ActionCodec}, a high-performance action tokenizer that instantiates the empirically validated design desiderata while incorporating the following practical enhancements.

\textbf{Embodiment-specific Soft-prompts.} To support multi-embodiment data integration and cross-platform knowledge transfer, we augment the Perceiver-based backbone with learnable \textit{soft-prompt} embeddings. While global network parameters are shared to distill universal robotic priors, distinct soft-prompts are assigned to each embodiment to capture unique mechanical constraints and control frequencies. This hierarchical representational strategy enables ActionCodec to undergo effective large-scale pre-training across diverse datasets, including LIBERO \citep{liu2024libero}, BridgeData \citep{walke2023bridgedata}, and DROID \citep{khazatsky2024droid}. With a unified action latent space, we facilitate zero-shot action re-targeting and accelerate fine-tuning on novel robotic platforms. (See soft-prompts details in \Cref{appendix:soft_prompt}.)

\textbf{RVQ Post-training.} While our analysis indicates that VLA optimization is robust to reconstruction errors below a critical threshold, maximizing fidelity is essential for fine-grained manipulation. However, standard Residual Vector Quantization (RVQ) \citep{lee2022rvq} typically achieves high fidelity at the expense of topological stability, often reducing the overlap rate (OR) to below $20\%$, which severely destabilizes VLA training. To resolve this trade-off, we propose an \textit{RVQ post-training} phase. We first pre-train a single-layer VQ model to maximize OR and perceptual alignment, establishing a stable supervision manifold. Subsequently, we freeze the encoder and the primary codebook, introducing additional residual codebooks to minimize the remaining reconstruction error. It allows ActionCodec to refine action precision orthogonally to the token's topological stability, ensuring zero degradation to VLA training efficiency while facilitating compatibility with advanced prediction paradigms such as Block-wise Autoregression (BAR) \citep{liu2025faster}. (See RVQ details in \Cref{appendix:rvq_ft}.)

\vspace{-5pt}
\section{Experiments}
We evaluate \textbf{ActionCodec} across diverse benchmarks to answer four questions: (i) whether it improves efficiency and success rate over heuristic and data-driven tokenizers; (ii) whether it benefits prevailing VLA paradigms; (iii) how it transfers to real robots; and (iv) which design choices matter.

\subsection{Comparison with Mainstream Action Tokenizers}

\begin{table}[t]
    \centering
    \caption{{\small\textbf{Benchmark Results.} Panels (a)--(c) report LIBERO success rates (\%) for pre-trained VLA backbones, scratch baselines, and SmolVLM2 tokenizer/variant comparisons, respectively. Panel (d) reports SimplerWidowX success rates (\%). \textbf{Pt.} denotes VLA pre-training of the policy backbone, excluding tokenizer pre-training. \textbf{Avg. Rank} indicates relative performance across tasks. Best results in each block are \textbf{bolded}, second best are \underline{underlined}.}}
    \vspace{5pt}
    \label{tab:libero_results_refined_final}
    \scriptsize
    \begin{minipage}[t]{0.49\linewidth}
    \centering
    \textbf{(a) LIBERO: Pre-trained VLA Backbones}
    \resizebox{\linewidth}{!}{%
    \begin{tabular}{l@{\hspace{4pt}}c@{\hspace{6pt}}c@{\hspace{10pt}}c@{\hspace{10pt}}c@{\hspace{10pt}}c@{\hspace{10pt}}c@{\hspace{10pt}}c}
    \toprule
    \textbf{Model} & \textbf{Pt.} & \textbf{Goal} & \textbf{Spatial} & \textbf{Object} & \textbf{Long} & \textbf{Avg.} & \makecell[c]{\textbf{Avg.}\\\textbf{Rank}} \\
    \midrule
    Octo-Base \citep{mees2024octo}            & \textcolor{gray}{Y} & 84.6 & 78.9 & 85.7 & 51.1 & 75.1 & 10.5 \\
    SpatialVLA \citep{qu2025spatialvla}       & \textcolor{gray}{Y} & 78.6 & 88.2 & 89.9 & 55.5 & 78.1 & 9.5 \\
    GR00T-N1 \citep{bjorck2025gr00t}          & \textcolor{gray}{Y} & 93.0 & 94.4 & 97.6 & 90.6 & 93.9 & 6.6 \\
    UniVLA \citep{bu2025univla}               & \textcolor{gray}{Y} & 95.6 & 96.5 & 96.8 & 92.0 & 95.2 & 5.4 \\
    VQ-VLA \citep{wang25vqvla}                & \textcolor{gray}{Y} & 75.2 & -    & -    & 60.0 & -    & - \\
    OpenVLA \citep{kim2024openvla}            & \textcolor{gray}{Y} & 79.2 & 84.7 & 88.4 & 53.7 & 76.5 & 10.0 \\
    OpenVLA-OFT \citep{kim2025openvlaoft}     & \textcolor{gray}{Y} & \underline{97.9} & \underline{97.6} & 98.4 & \underline{94.5} & \underline{97.1} & \underline{2.5} \\
    $\pi_0$ \citep{black2024pi0}              & \textcolor{gray}{Y} & 95.8 & 96.8 & 98.8 & 85.2 & 94.2 & 4.5 \\
    $\pi_{0.5}$ \citep{intelligence2025pi05}  & \textcolor{gray}{Y} & \textbf{98.0} & \textbf{98.8} & 98.2 & 92.4 & 96.8 & \underline{2.5} \\
    $\pi_0$ FAST \citep{pertsch2025fast}      & \textcolor{gray}{Y} & 88.6 & 96.4 & 96.8 & 60.2 & 85.5 & 7.4 \\
    \textcolor{gray}{ActionCodec (2.2B)}      & \textcolor{gray}{N} & \textcolor{gray}{95.4} & \textcolor{gray}{96.2} & \textcolor{gray}{\underline{99.6}} & \textcolor{gray}{90.6} & \textcolor{gray}{95.5} & \textcolor{gray}{5.1} \\
    \textcolor{gray}{ActionCodec-BAR}         & \textcolor{gray}{N} & \textcolor{gray}{97.6} & \textcolor{gray}{97.2} & \textcolor{gray}{\textbf{99.8}} & \textcolor{gray}{\textbf{94.8}} & \textcolor{gray}{\textbf{97.4}} & \textcolor{gray}{\textbf{2.0}} \\
    \bottomrule
    \end{tabular}}

    \vspace{4pt}

    \textbf{(b) LIBERO: Scratch Baselines}
    \resizebox{\linewidth}{!}{%
    \begin{tabular}{l@{\hspace{4pt}}c@{\hspace{6pt}}c@{\hspace{10pt}}c@{\hspace{10pt}}c@{\hspace{10pt}}c@{\hspace{10pt}}c@{\hspace{10pt}}c}
    \toprule
    \textbf{Model} & \textbf{Pt.} & \textbf{Goal} & \textbf{Spatial} & \textbf{Object} & \textbf{Long} & \textbf{Avg.} & \makecell[c]{\textbf{Avg.}\\\textbf{Rank}} \\
    \midrule
    Diffusion Policy \citep{chi2023diffusionpolicy}& \textcolor{gray}{N} & 68.3 & 78.3 & 92.5 & 50.5 & 72.4 & 13.8 \\
    MiniVLA \citep{belkhale2024minivla}       & \textcolor{gray}{N} & -    & -    & -    & 77.0 & -    & - \\
    OpenVLA-OFT \citep{kim2025openvlaoft}     & \textcolor{gray}{N} & 91.7 & 94.3 & 95.2 & 86.5 & 91.9 & 7.8 \\
    SmolVLA 2.2B \citep{shukor2025smolvla}    & \textcolor{gray}{N} & 91.0 & 93.0 & 94.0 & 77.0 & 88.8 & 10.2 \\
    $\pi_0$ FAST \citep{pertsch2025fast}      & \textcolor{gray}{N} & 89.0 & 87.0 & 63.0 & 48.0 & 71.8 & 14.2 \\
    $\pi_{0.5}$ \citep{intelligence2025pi05}  & \textcolor{gray}{N} & 94.6 & 96.6 & 97.2 & 85.8 & 93.3 & 6.4 \\
    VLA-0 \citep{goyal2025vla0}               & \textcolor{gray}{N} & 96.2 & \underline{97.0} & 97.8 & 87.6 & 94.7 & 3.8 \\
    \bottomrule
    \end{tabular}}
    \end{minipage}\hfill
    \begin{minipage}[t]{0.49\linewidth}
    \centering
    \textbf{(c) LIBERO: SmolVLM2 + Tokenizers and Variants}
    \resizebox{\linewidth}{!}{%
    \begin{tabular}{l@{\hspace{4pt}}c@{\hspace{6pt}}c@{\hspace{10pt}}c@{\hspace{10pt}}c@{\hspace{10pt}}c@{\hspace{10pt}}c@{\hspace{10pt}}c}
    \toprule
    \textbf{Model} & \textbf{Pt.} & \textbf{Goal} & \textbf{Spatial} & \textbf{Object} & \textbf{Long} & \textbf{Avg.} & \makecell[c]{\textbf{Avg.}\\\textbf{Rank}} \\
    \midrule
    \rowcolor[HTML]{F2F2F2}
    \multicolumn{8}{l}{\textit{SmolVLM2-2.2B (256M/500M) + Tokenizers}} \\
    Binning                                   & \textcolor{gray}{N} & 27.4 & 64.2 & 71.0 & 50.8 & 53.4 & 15.5 \\
    String                                    & \textcolor{gray}{N} & 30.8 & 69.6 & 71.6 & 26.4 & 49.6 & 15.5 \\
    VQVLA's                                   & \textcolor{gray}{N} & 60.5 & 56.0 & 69.8 & 51.4 & 60.5 & 15.2 \\
    MiniVLA's                                 & \textcolor{gray}{N} & 82.1 & 82.8 & 90.2 & 75.2 & 82.6 & 12.8 \\
    FAST                                      & \textcolor{gray}{N} & 89.2 & 90.2 & 97.2 & 85.8 & 90.6 & 9.4 \\
    ActionCodec \textcolor{gray}{(256M)}      & \textcolor{gray}{N} & 91.2 & 94.2 & 97.0 & 86.8 & 92.3 & 7.5 \\
    ActionCodec \textcolor{gray}{(500M)}      & \textcolor{gray}{N} & 96.4 & 91.6 & 96.0 & 85.8 & 92.5 & 7.8 \\
    ActionCodec (2.2B)                        & \textcolor{gray}{N} & 95.4 & 96.2 & \underline{99.6} & 90.6 & 95.5 & 3.8 \\
    \rowcolor[HTML]{F2F2F2}
    \multicolumn{8}{l}{\textit{Variants}} \\
    ActionCodec-PD                            & \textcolor{gray}{N} & \underline{97.0} & 95.2 & 98.4 & \underline{91.6} & \underline{95.6} & \underline{3.0} \\
    ActionCodec-KI                            & \textcolor{gray}{N} & 95.0 & 93.6 & 97.6 & 90.8 & 94.3 & 5.5 \\
    ActionCodec-BAR                           & \textcolor{gray}{N} & \textbf{97.6} & \textbf{97.2} & \textbf{99.8} & \textbf{94.8} & \textbf{97.4} & \textbf{1.0} \\
    \bottomrule
    \end{tabular}}

    \vspace{4pt}

    \textbf{(d) SimplerWidowX Benchmark Results}
    \resizebox{\linewidth}{!}{%
    \begin{tabular}{l@{\hspace{4pt}}c@{\hspace{10pt}}c@{\hspace{10pt}}c@{\hspace{10pt}}c@{\hspace{10pt}}c@{\hspace{10pt}}c@{\hspace{10pt}}c}
    \toprule
    \textbf{Model} & \textbf{Pt.} & \textbf{Spoon} & \textbf{Carrot} & \textbf{Block} & \textbf{Eggplant} & \textbf{Avg.} & \makecell[c]{\textbf{Avg.}\\\textbf{Rank}} \\
    \midrule\midrule
    Octo-Base \citep{mees2024octo}            & \textcolor{gray}{Y} & 12.5 & 8.3 & 0.0 & 43.1 & 16.0 & 7.3 \\
    SpatialVLA \citep{qu2025spatialvla}       & \textcolor{gray}{Y} & 16.7 & 25.0 & 29.2 & \textbf{100.0} & 42.7 & 4.0 \\
    $\pi_0$ \citep{black2024pi0}              & \textcolor{gray}{Y} & \underline{66.7} & 58.3 & \textbf{58.3} & \underline{88.3} & \textbf{67.9} & \textbf{2.0} \\
    OpenVLA-OFT \citep{kim2025openvlaoft}     & \textcolor{gray}{Y} & 12.5 & 4.2 & 8.3 & 0.0 & 6.3 & 8.1 \\
    UniVLA \citep{bu2025univla}               & \textcolor{gray}{Y} & 54.2 & \textbf{66.7} & 50.0 & 4.2 & 43.8 & \underline{3.5} \\
    OpenVLA \citep{kim2024openvla}            & \textcolor{gray}{Y} & 32.0 & 30.0 & 18.0 & 38.0 & 29.5 & 4.8 \\
    VQ-VLA \citep{wang25vqvla}                & \textcolor{gray}{Y} & 12.5 & 8.0 & 6.0 & 0.0 & 6.6 & 8.1 \\
    $\pi_0$ FAST \citep{pertsch2025fast}      & \textcolor{gray}{Y} & 29.1 & 21.9 & 10.8 & 66.6 & 32.1 & 5.3 \\
    \textcolor{gray}{ActionCodec-BAR}         & \textcolor{gray}{N} & \textcolor{gray}{\textbf{71.7}} & \textcolor{gray}{\underline{64.2}} & \textcolor{gray}{\underline{55.0}} & \textcolor{gray}{70.0} & \textcolor{gray}{\underline{65.2}} & \textcolor{gray}{\textbf{2.0}} \\
    \midrule
    MiniVLA \citep{belkhale2024minivla}       & \textcolor{gray}{N} & \underline{68.0} & \underline{44.0} & \textbf{70.0} & \underline{14.0} & \underline{49.0} & \underline{1.8} \\
    ActionCodec-BAR                           & \textcolor{gray}{N} & \textbf{71.7} & \textbf{64.2} & \underline{55.0} & \textbf{70.0} & \textbf{65.2} & \textbf{1.3} \\
    \bottomrule
    \end{tabular}}
    \end{minipage}
    \vspace{-12pt}
\end{table}

For a controlled comparison, we plug all tokenizers into the same SmolVLM2-2.2B \citep{marafioti2025smolvlm} VLA by vocabulary expansion and full-parameter autoregressive fine-tuning on LIBERO; prompts appear in \Cref{fig:vlm_prompt}, and comparison protocols in \Cref{appendix:fairness_efficiency}. Baselines include \textbf{Binning} with 1,000 bins per action dimension \citep{goyal2025vla0}, raw-action \textbf{String} tokens \citep{hancock2025vlm2vla}, VQ-based MiniVLA \citep{belkhale2024minivla} and VQ-VLA \citep{wang25vqvla}, and \textbf{FAST} \citep{pertsch2025fast}. As shown by the learning curves in \Cref{fig:teaser}, ActionCodec reaches both the best peak success rate and the fastest convergence: 89.5\% at 5K steps versus 38.6\% for the runner-up FAST. Panel (c) of \Cref{tab:libero_results_refined_final} further shows that ActionCodec with 256M/500M VLMs outperforms other tokenizers using a 2.2B backbone, indicating that action representation quality is as important as VLM scale.

We also report throughput and latency in \Cref{tab:tokenizer_efficiency}: Binning and String incur prohibitive latency and low throughput for real-time control, whereas ActionCodec achieves the highest action throughput while preserving the best task performance.

\begin{wraptable}[11]{r}{0.58\textwidth}
    \vspace{-10pt}
    \centering
    \caption{{\small \textbf{Action Tokenization Performance and Efficiency.} Comparison of token budget, overlap rate (OR), and inference metrics. All models are evaluated on the LIBERO benchmark.}}
    \label{tab:tokenizer_efficiency}
    \scriptsize
    \resizebox{\linewidth}{!}{%
    \begin{tabular}{lcccc|cc}
    \toprule
    \textbf{Tokenizer} & \makecell[c]{\textbf{Token}\\\textbf{Budget}} & \textbf{OR} & \textbf{Horizon} & \makecell[c]{\textbf{Latency}\\\textbf{(s)}} & \makecell[c]{\textbf{Throughput}\\\textbf{(action/s)}} & \textbf{SR (\%)} \\
    \midrule
    Binning \citep{goyal2025vla0}        & 140.0±0.0  & 39\% & 20 & 4.9  & 4.1  & 3.5  \\
    Binning \citep{goyal2025vla0}        & 56.0±0.0   & 38\% & 8  & 2.6  & 3.1  & 53.4 \\
    String \citep{hancock2025vlm2vla}  & 779.2±17.3 & 28\% & 20 & 33.8 & 0.6  & 20.4 \\
    String \citep{hancock2025vlm2vla}  & 311.8±7.8  & 34\% & 8  & 12.7 & 0.6  & 49.6 \\
    VQVLA's \citep{wang25vqvla} & 4.0±0.0    & 45\% & 5  & 0.7  & 7.7  & 60.5 \\
    MiniVLA's \citep{belkhale2024minivla} & 7.0±0.0    & 40\% & 8  & 0.7  & 11.1 & 82.6 \\
    FAST \citep{pertsch2025fast} & 24.3±9.1   & 19\% & 20 & 1.5  & 13.1 & 90.6 \\
    {ActionCodec} & {16.0±0.0} & {72\%} & {20} & {0.9} & \textbf{22.0} & \textbf{95.5} \\
    \bottomrule
    \end{tabular}
    }
    \vspace{-12pt}
\end{wraptable}

\subsection{Integration with Prevailing VLA Paradigms}\label{sec:paradigms_exp}

To test architectural versatility, we combine ActionCodec with three VLA paradigms: \textbf{Parallel Decoding (PD)}, which predicts tokens in one bidirectional action-expert pass \citep{kim2025openvlaoft}; \textbf{Knowledge Isolation (KI)}, which uses a diffusion action expert conditioned on the VLM KV cache to decouple semantics from action generation \citep{intelligence2025pi05}; and \textbf{Block-wise Autoregression (BAR)}, which predicts RVQ codebook levels in parallel while preserving inter-block autoregressive dependencies \citep{liu2025faster}. Details are in \Cref{appendix:paradigm}.

Panels (a)--(c) of \Cref{tab:libero_results_refined_final} show consistent gains across paradigms. PD and BAR outperform native autoregression, and \textbf{ActionCodec-BAR} reaches 97.4\% average success on LIBERO, setting a new SOTA for VLAs without VLA pre-training and surpassing several VLA-pre-trained models. KI is slightly below the autoregressive ActionCodec baseline but still exceeds its FAST-based counterpart, represented by $\pi_{0.5}$ without VLA pre-training. On the OOD Simpler-WidowX benchmark (panel (d)), ActionCodec-BAR keeps the best rank among models with or without prior robotic data. Additional large-scale VLA pre-training and transfer results are in \Cref{appendix:large_scale}.

\begin{figure}[t]
    \centering
    \includegraphics[width=0.9\linewidth]{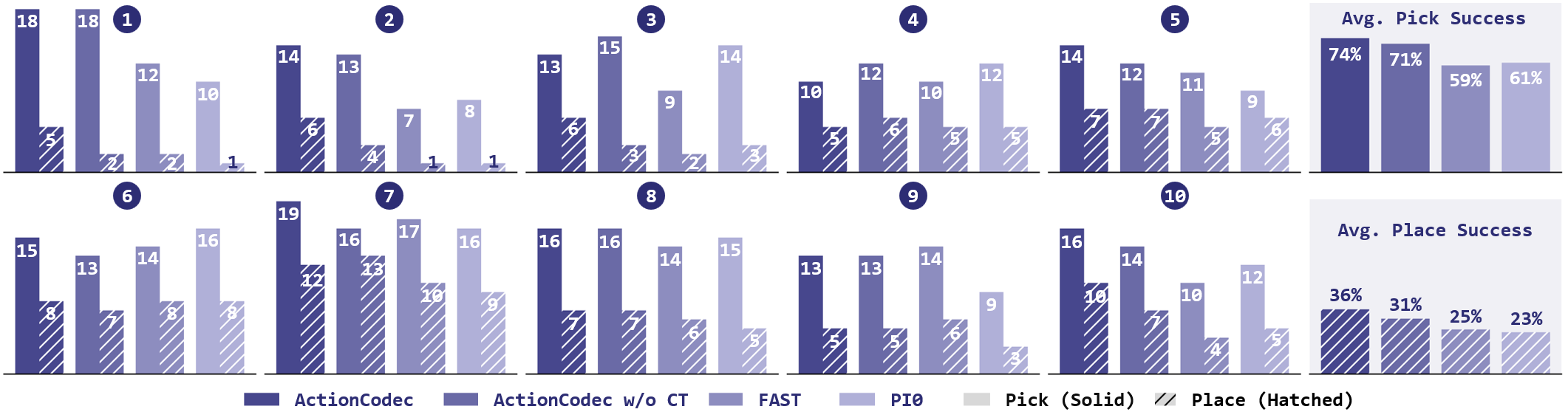}
    \caption{\textbf{Success Rates on SO100-ShapeSorter.} We report success counts for \textit{Pick} (solid) and \textit{Place} (hatched) across 10 tasks. While recovery actions are inherently present in our task demonstrations due to the difficulty of slot insertion, only \textbf{ActionCodec} (with CT) successfully learns to reproduce these corrective strategies. The performance gap is most prominent in the \textit{Place} phase, where co-training facilitates the execution of complex recovery behaviors that task-only models fail to capture.}
    \label{fig:so100_bar_chart}
    \vspace{-10pt}
\end{figure}

\subsection{Real-world Evaluation}

\begin{wrapfigure}{r}{0.5\textwidth}
    \vspace{-24pt}
    \centering
    \includegraphics[width=\linewidth]{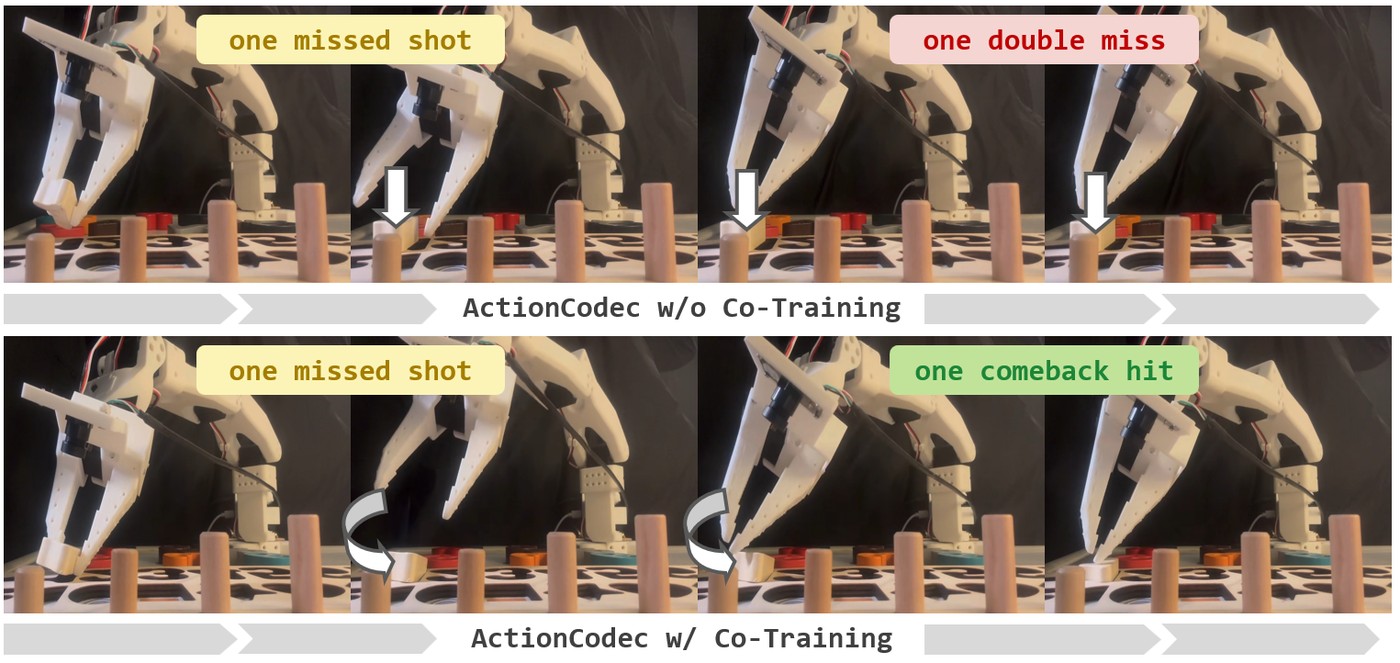}
    \caption{{\small\textbf{Comparison of recovery behaviors on SO100.} After an initial missed shot, {ActionCodec} with co-training successfully executes a corrective adjustment to complete the insertion, whereas the variant without co-training stalls after a failure.}}
    \label{fig:so100_inference_demo}
    \vspace{-5pt}
\end{wrapfigure}

\textbf{SO100-ShapeSorter.}\label{sec:so100_exp} SO100-ShapeSorter\footnote{SO100 Dataset: \url{https://huggingface.co/datasets/ZibinDong/actioncodec-so100-part1}} (\Cref{fig:benchmarks}) evaluates ActionCodec on a low-cost robot and tests large-scale data integration. The SO100 arm must insert numbered blocks into matching slots; the narrow tolerance makes insertion difficult even for humans, often requiring corrective attempts, so the task dataset naturally contains stochastic failure-and-readjustment \textit{recovery behaviors}. We isolate co-training (CT) by comparing ActionCodec trained with 22.9K SO100 community episodes \citep{shukor2025smolvla} to a task-only \textbf{ActionCodec w/o CT}, plus $\pi_0$ and $\pi_0$ FAST. In \Cref{fig:so100_bar_chart}, all models pick reasonably, but \textit{Place} reveals the key gap: despite recovery trajectories in the task demonstrations, \textit{w/o CT} often stalls after initial misalignment, whereas co-trained ActionCodec uses broader priors to model corrective actions and recover in closed loop, e.g., by re-aligning after a failed insertion (\Cref{fig:so100_inference_demo}). Thus, small task datasets may contain recovery data, but diverse co-training priors are needed to internalize and execute these long-tail behaviors.

\textbf{xArm-PickVeg.}\label{sec:xarm_exp} xArm-PickVeg\footnote{xArm Dataset: \url{https://huggingface.co/datasets/ZibinDong/xarm_pickplace}} (\Cref{fig:benchmarks}) tests ActionCodec on a high-performance platform and measures representational pre-training transfer. We compare \textbf{w/ PT}, initialized from large-scale heterogeneous pre-training, with \textbf{w/o PT}, randomly initialized. In \Cref{fig:xarm_curves}, \textit{w/ PT} optimizes better, with lower $L_2$ reconstruction error and higher OR than \textit{w/o PT}; this stability improves control, yielding an 82.5\% success rate in \Cref{tab:xarm_results}. ActionCodec therefore supports efficient fine-tuning and robust transfer to novel high-performance robots.

\begin{table}[t]
    \centering
    \vspace{-5pt}
    \begin{minipage}[t]{0.48\textwidth}
        \centering
        \caption{\textbf{Success Rates on xArm-PickVeg.} ActionCodec w/ PT leverages heterogeneous priors to achieve the highest success rate.}
        \label{tab:xarm_results}
        \scriptsize
        \resizebox{\linewidth}{!}{%
            \begin{tabular}{l|cccc}
                \toprule
                \textbf{Model} & \makecell[c]{\textbf{ActionCodec} \\ \textbf{w/ PT}} & \makecell[c]{\textbf{ActionCodec} \\ \textbf{w/o PT}} & ${\pi_0}$ & ${\pi_0}$ \textbf{FAST} \\
                \midrule
                \textbf{SR (\%)} & \textbf{82.5} & 74.1 & 75.0 & 72.5 \\
                \bottomrule
            \end{tabular}
        }
    \end{minipage}\hfill
    \begin{minipage}[t]{0.48\textwidth}
        \centering
        \caption{\textbf{Ablations.} We ablate various design choices on LIBERO. We report the average success rate over the four suites.}
        \label{tab:ablation_study}
        \newcommand{\bluecheck}{\textcolor{green!70!black}{\scalebox{1.2}{$\bm\checkmark$}}}
        \newcommand{\redcross}{\textcolor{red}{\scalebox{1.2}{$\bm\times$}}}
        \newcommand{\gain}[1]{\textsuperscript{\textcolor{green!70!black}{(+#1\%)}}}
        \newcommand{\loss}[1]{\textsuperscript{\textcolor{red}{(-#1\%)}}}
        \scriptsize
        \resizebox{\linewidth}{!}{%
        \begin{tabular}{l ccc c l}
        \toprule
        \textbf{ID} & \textbf{Co-Training} & \textbf{Soft Prompt} & \textbf{RVQ Post Training} & \textbf{VLM} & \textbf{LIBERO Avg.} \\
        \midrule
        0 & \redcross & \redcross & \bluecheck & \raisebox{-0.8ex}{\includegraphics[height=3ex]{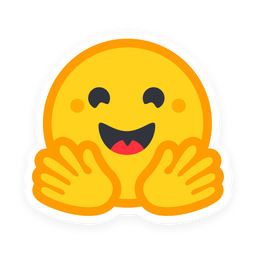}} & 92.0 \loss{3.5} \\
        1 & \bluecheck & \redcross & \bluecheck & \raisebox{-0.8ex}{\includegraphics[height=3ex]{figures/hf-logo.png}} & 92.7 \loss{2.8} \\
        2 & \bluecheck & \bluecheck & \redcross & \raisebox{-0.8ex}{\includegraphics[height=3ex]{figures/hf-logo.png}} & 95.2 \loss{0.3} \\
        \rowcolor[HTML]{F2F2F2}
        3 & \bluecheck & \bluecheck & \bluecheck & \raisebox{-0.8ex}{\includegraphics[height=3ex]{figures/hf-logo.png}} & \textbf{95.5} \\
        \midrule
        4 & \bluecheck & \bluecheck & \bluecheck & \raisebox{-0.8ex}{\includegraphics[height=3ex]{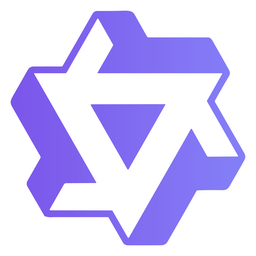}} & 95.1 \\
        5 & \bluecheck & \bluecheck & \bluecheck & \raisebox{-0.8ex}{\includegraphics[height=3ex]{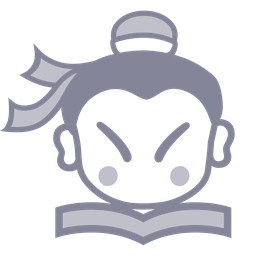}} & 94.6 \\
        \bottomrule
        \end{tabular}
        }
    \end{minipage}
    \vspace{-15pt}
\end{table}

\subsection{Ablation Study}

We ablate ActionCodec on LIBERO in \Cref{tab:ablation_study}. \textbf{Soft Prompts.} Removing embodiment-specific soft prompts (ID 1) reduces the gains from diverse pre-training, while Bridge-WidowX-to-LIBERO-Franka/DROID-Franka/xArm transfers (\Cref{fig:action_transfer}) show congruent motions, indicating a unified, platform-agnostic latent space.\textbf{~RVQ Post-training.} RVQ post-training lowers reconstruction error but only modestly affects success rate (ID 2 vs. 3), confirming that topological stability (OR) matters more than absolute fidelity for VLA optimization.\textbf{~VLM Compatibility.} ActionCodec remains backbone-agnostic, with robust and comparable performance across IDs 3--5:
\raisebox{-0.5ex}{\includegraphics[height=3ex]{figures/hf-logo.png}} SmolVLM2-2.2B \citep{marafioti2025smolvlm}, 
\raisebox{-0.5ex}{\includegraphics[height=3ex]{figures/qwen-color.png}} Qwen2.5VL-3B \citep{qwen25}, and 
\raisebox{-0.5ex}{\includegraphics[height=3ex]{figures/internlm-color.png}} InternVL3.5-2B \citep{wang2025internvl3}. 

\section{Conclusion}

In this paper, we systematically analyze the critical design elements of action tokenization, leading to the introduction of \textbf{ActionCodec}. By integrating identified best practices, ActionCodec achieves superior performance across diverse simulation and real-world benchmarks. Our additional large-scale results (\Cref{appendix:large_scale}) support transfer, while robust in-the-wild generalization across broader embodiments remains future work. Furthermore, investigating optimized neural architectures and refined vision-language alignment schemes remains a promising direction. We believe ActionCodec can provide a clear roadmap for the community to develop next-generation physical intelligence.

\section{Limitations}

ActionCodec has three main limitations. First, the current neural architecture is chosen for flexible design-choice analysis rather than optimized for final performance. Second, training relies on two contrastive auxiliary losses, which introduce additional hyperparameters and make tuning more involved. Third, the additional 15K-hour, 14-embodiment study (\Cref{appendix:large_scale}) covers a limited set of benchmarks; broad in-the-wild generalization remains to be established.


\section*{Impact Statement}
This paper presents work whose goal is to advance the field of Machine
Learning. There are many potential societal consequences of our work, none
which we feel must be specifically highlighted here.



\bibliography{example_paper}

\newpage
\appendix

\onecolumn

\section{Related Works}

\textbf{Vision-Language-Action Models.} VLAs leverage the representational power of pre-trained VLMs to map visual observations and language instructions into robotic control signals \citep{driess2023palm, zitkovich2023rt2}. Current VLA architectures primarily diverge in their action representation, falling into either continuous or discrete prediction frameworks. Continuous frameworks often necessitate structural deviations from the VLM's native architecture, such as regressing actions from hidden states or integrating auxiliary generative experts such as diffusion or flow-matching models \citep{black2024pi0, shukor2025smolvla, bjorck2025gr00t, kim2025openvlaoft, yuan2026fastwam}. Conversely, discrete frameworks discretize action spaces into tokens, enabling the model to utilize the standard autoregressive objective \citep{wang25vqvla, goyal2025vla0, liu2026g05}. While discrete methods may introduce higher sequential latency, they demonstrate a superior ability to preserve the VLM’s world knowledge and instruction-following capabilities \citep{liu2025faster}. Notably, even SOTA heterogeneous models have recently integrated discrete action tokenization as an auxiliary task to improve performance \citep{driess2025KI, intelligence2025pi05}. This convergence underscores the pivotal role of action tokenization in the VLA design space.

\textbf{Action Tokenization Schemes.} The methodology for action discretization varies significantly. Early approaches employed uniform quantization (binning) of temporal signals \citep{kim2024openvla, lee2025molmoact}, which often suffers from low training efficiency and ignores the inherent structure of the action space. Subsequent refinements, such as parallel decoding \citep{kim2025openvlaoft} or string-based token representation \citep{goyal2025vla0}, attempt to mitigate these issues but do not address the fundamental inefficiencies of heuristic binning. Other studies have attempted to represent actions as strings for direct VLM prediction \citep{hancock2025vlm2vla}; however, this approach offers no significant performance benefits while greatly increasing the token budget and extending latency to several seconds, limiting real-world applicability. FAST \citep{pertsch2025fast} introduces Byte-Pair Encoding (BPE) on frequency-domain signals; however, its reliance on fixed geometric priors limits its capacity for cross-embodiment knowledge transfer. Data-driven schemes, particularly those based on Vector Quantization (VQ) \citep{wang25vqvla, belkhale2024minivla, mete2024quest, lee2024vqbet}, offer a more flexible alternative by learning discrete latent representations. Despite their potential, existing VQ-based tokenizers are often treated as black-box components. There remains a critical lack of understanding regarding the specific VQ properties that facilitate or obstruct VLA optimization, a gap that this work aims to bridge.

\begin{figure}[t]
    \centering
    \includegraphics[width=1.0\linewidth]{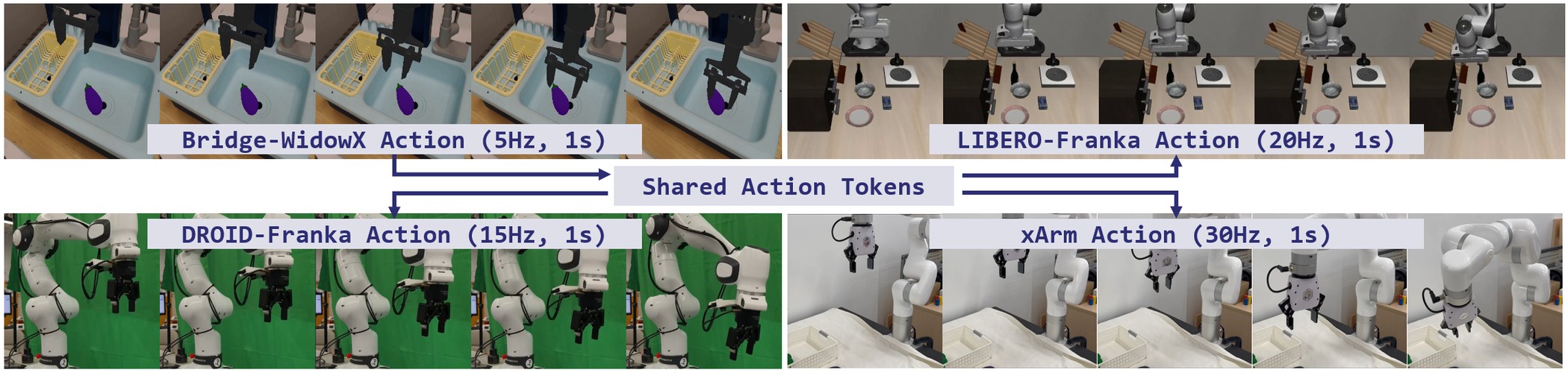}
    \caption{\textbf{Cross-embodiment action transfer.} To evaluate the transferability of ActionCodec, a 1-second WidowX action sequence sampled from BridgeData is encoded into action tokens and subsequently decoded into 1-second trajectories for LIBERO-Franka, DROID-Franka, and xArm. After deploying these sequences twice (totaling 2 seconds) and recording the resulting motions, all three robotic platforms exhibit highly consistent action patterns. This demonstrates that ActionCodec effectively captures hardware-agnostic action semantics.}
    \label{fig:action_transfer}
\end{figure}

\section{Experimental Setup and Benchmark Details}

\subsection{Benchmarks}

\begin{figure}
    \vspace{-10pt}
    \centering
    \includegraphics[width=0.5\linewidth]{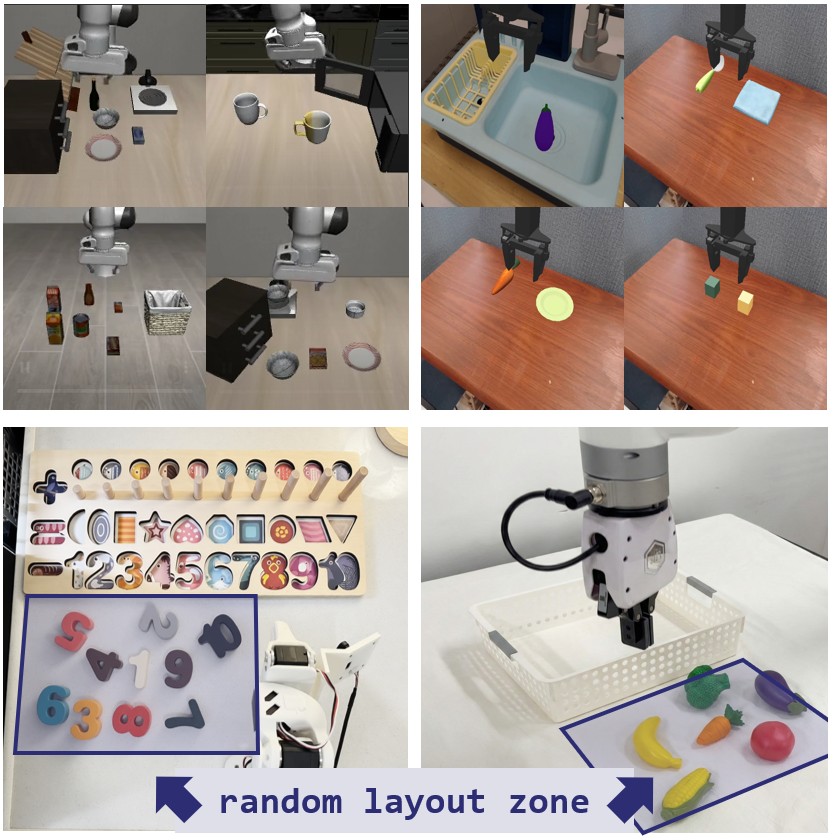}
    \caption{{\small \textbf{All evaluation benchmarks.} We evaluate VLA models on 4 LIBERO task suites, Bridge-WidowX tasks on SimplerEnv, SO100-ShapeSorter tasks, and xArm tabletop manipulation tasks.}}
    \label{fig:benchmarks}
    \vspace{-12pt}
\end{figure}

\textbf{LIBERO.} The LIBERO benchmark \citep{liu2024libero} comprises four task suites: \textit{Goal}, \textit{Spatial}, \textit{Object}, and \textit{Long}. Each suite contains 10 tasks, with 50 demonstrations provided per task. These suites evaluate the model's performance in instruction following, spatial reasoning, object recognition, and long-horizon task completion. Following the official $\pi_0$ fine-tuning recipe, we fine-tune the VLM under a multi-task joint training setting for 30k steps. Evaluation is conducted over 50 trials per task using the predefined initial states.

\textbf{SimplerEnv-WidowX.} To assess zero-shot generalization, we utilize SimplerEnv-WidowX \cite{li2024simplerenv}, which provides four out-of-domain desktop manipulation tasks not included in the BridgeData set. We fine-tune the VLM on BridgeData for 4 epochs before direct evaluation. To mitigate the high variance typical of zero-shot evaluation, we increase the number of trials per task from the default 24 to 120.

\textbf{SO100-ShapeSorter.} We conduct real-world evaluations using the SO100 platform in a shape-sorting task. The robot must identify a specific numerical block (1–10) based on language instructions and insert it into the matching slot. The SO100 robotic arms were assembled according to the official LeRobot \citep{cadene2024lerobot} documentation\footnote{Official SO100 document: \url{https://huggingface.co/docs/lerobot/so100}}. Two 480x640 RGB cameras are utilized, mounted on the gripper and directly overhead, respectively. Data collection is performed at a frequency of 30Hz. Both proprioception and action spaces are defined by absolute joint angles. This benchmark tests both semantic recognition and fine-grained manipulation. We collect 50 demonstrations per block and fine-tune the VLM for 30k steps. During testing, block positions are randomized, and 20 trials are performed for each number. We record task progress at two checkpoints: 50\% for a successful grasp and 100\% for successful insertion. In \Cref{sec:so100_exp}, we distinguish between two variants: \textit{w/ co-training} refers to ActionCodec trained on LIBERO \citep{liu2024libero}, BridgeData \citep{walke2023bridgedata}, DROID \citep{khazatsky2024droid}, the SO100-community dataset \citep{shukor2025smolvla}, and the SO100-ShapeSorter dataset; \textit{w/o co-training} excludes the SO100-community dataset. Detailed \texttt{repo\_ids} for the community dataset are available in \citet{shukor2025smolvla} Appendix A.1.

\textbf{xArm-PickVeg.} This benchmark involves a multi-task vegetable picking task where the xArm robot identifies and grasps a target vegetable model based on language instructions. The system uses only RGB images from a third-person Intel Realsense L515 camera, with data acquisition at 30Hz. Proprioception utilizes the end-effector (EEF) pose, while the action space is defined by delta EEF poses. We collect 30 demonstrations for each vegetable and fine-tune the VLM for 30k steps. During testing, object positions are randomized, and 20 trials are performed for each item. We utilize a binary success metric, as most failures are concentrated in the recognition or initial grasping phase. In \Cref{sec:xarm_exp}, we denote the variant initialized with weights pre-trained on LIBERO, BridgeData, and DROID as \textit{w/ PT}, while \textit{w/o PT} refers to a randomly initialized model trained solely on the xArm dataset.

\subsection{Details of Validation Experiments}\label{appendix:validation_exp}
To rigorously evaluate the design desiderata discussed in \Cref{sec:before_method}, we provide comprehensive details of our validation experimental setup.

\textbf{VQ Tokenizer Architecture.} As illustrated in \Cref{fig:arch}, we implement the encoder and decoder using a Perceiver-based Transformer architecture \citep{jaegle2021perceiver}. This design provides the structural flexibility required for our ablation studies: utilizing only cross-attention layers removes direct interaction between latent queries, without implying statistical independence of the resulting tokens, while the integration of an auxiliary self-attention layer (with either bidirectional or causal masking) allows for the precise modulation of inter-token dependencies. This modularity is essential for our analysis of \textit{residual grammar} in \Cref{sec:vl_align_and_grammar}. While alternative backbones such as CNNs \citep{lee2024vqbet, belkhale2024minivla, wang25vqvla} or standard Transformers \citep{liu2025faster} are prevalent in literature, we treat the specific choice of backbone as an orthogonal engineering consideration and focus primarily on the inherent properties of the tokenization scheme.

\textbf{VLA Implementation.} To isolate the influence of the action tokenizer, we adhere to a native VLM-style autoregressive paradigm. We perform full-parameter fine-tuning on SmolVLM2-256M \citep{marafioti2025smolvlm} to predict action tokens without any robotics-specific architectural modifications (\Cref{tab:vla_hyperparams_validation}). This model scale is chosen for its competitive multimodal performance and computational tractability for extensive ablation studies. Specifically, we expand the VLM vocabulary with $S$ special tokens: $\texttt{<|action\_0|>} \dots \texttt{<|action\_(S-1)|>}$. During inference, any invalid token sequence produced by the VLM is substituted with a zero-action chunk to maintain control stability. For clarity, the \texttt{[BOS]} notation used in our analysis refers to the context token immediately preceding the action sequence in the prompt (see \Cref{fig:vlm_prompt}), serving as a functional beginning-of-sentence marker for the action expert.

Notably, the \textbf{FAST tokenizer} \citep{pertsch2025fast} requires a specialized sequence-length alignment heuristic. Since FAST necessitates that the BPE-decoded sequence matches a specific length for valid reconstruction—a constraint frequently violated by autoregressive VLMs—we implement a soft-decoding approach involving zero-padding or truncation to the required length. The repair applies only to invalid sequence lengths and leaves valid FAST outputs unchanged. Without this adjustment, models trained with FAST exhibit negligible success rates throughout the majority of the training phase. 

\textbf{Datasets and Benchmarking.} While the action tokenizers are pre-trained on the comprehensive LIBERO dataset \citep{liu2024libero}, VLA fine-tuning and evaluation are conducted specifically on the LIBERO-Goal task suite. We monitor performance at 500, 1k, 5k, 10k, and 20k training steps. We observe that performance below 500 steps is indistinguishable across variants, while models typically reach asymptotic convergence by 20k steps, providing a complete window into the training dynamics of each tokenization scheme.

\subsubsection{Impacts of Overlap Rate}\label{appendix:or_controls}

In \Cref{sec:impacts_or}, we synthesize three tokenizer variants with controlled OR levels: 26\%, 40\%, and 70\%. This controlled variance is achieved by augmenting the standard VQ-VAE objective function with an auxiliary InfoNCE contrastive loss \citep{oord2019infonce}: $\mathcal{L}_\text{VQ} + \lambda\cdot\mathcal{L}_\text{InfoNCE},$ where $\lambda$ is a weighting hyperparameter that modulates the degree of latent clustering. For a given action sample $A$, we define a positive pair as the latent representation $Z = \mathcal{F}(A)$ and the representation of its perturbed version $Z^+ = \mathcal{F}(A + \eta), ~\eta \sim \mathcal{N}(0, \sigma^2)$. All other action latents within the same mini-batch serve as negative samples $Z^-$. By minimizing the distance between $Z$ and $Z^+$ in the latent manifold, we implicitly increase the probability that semantically similar actions are mapped to the same discrete code, thereby elevating the OR. Our experiments reveal that a higher overlap rate significantly enhances the training efficiency of the VLA model. However, in subsequent validation experiments, we observe that incorporating a \textit{Time Contrastive Loss} (TCL) or a \textit{CLIP-style} cross-modal loss (\Cref{sec:vl_align_and_grammar}) naturally yields higher OR values while providing superior training stability compared to the explicit InfoNCE-based action perturbation. Consequently, our final architecture favors these more stable contrastive objectives.

\paragraph{Definition and controls.}
Let $C_t$ and $C_{t+1}$ be the $n$-token sequences for adjacent action chunks $A_t=(a_t,\ldots,a_{t+T-1})$ and $A_{t+1}=(a_{t+1},\ldots,a_{t+T})$. We define
\begin{equation}\label{eq:overlap_rate}
    \mathrm{OR}=\mathbb{E}_{t}\!\left[1-\frac{d_H(C_t,C_{t+1})}{n}\right],
\end{equation}
where $d_H$ is the Hamming distance and the expectation averages adjacent pairs. Thus OR measures token consistency at corresponding positions, rather than overlap of token sets. In the controlled OR experiment, token budget ($n=16$), vocabulary ($S=2048$), action horizon, chunk stride, and downstream VLA are fixed; only contrastive strength is varied. This holds the motion smoothness and overlap protocol constant across variants. The $n,S$ sweeps separately maintain OR at 40\% by adjusting the contrastive weight; each setting's reconstruction error is shown in \Cref{fig:toy_example_ablation}. These controls study properties within the VQ family; modifying the official FAST tokenizer to enforce a chosen OR would change the baseline method.

\begin{table}[htbp]
    \centering
    \caption{\textbf{Codebook usage in the controlled OR experiment.} All variants use $n=16$ and $S=2048$. Utilization is measured on the validation set; PPL is the marginal token-usage perplexity $\exp(H(c))$, with entropy in nats. Reconstruction errors are from \Cref{fig:toy_example_ablation}.}
    \label{tab:or_codebook_usage}
    \small
    \begin{tabular}{ccccc}
        \toprule
        OR (\%) & Codebook utilization (\%) & PPL & $L_1$ error & SR at 500 steps (\%) \\
        \midrule
        26 & 100 & 2016 & 0.020 & 8.2 \\
        40 & 100 & 1583 & 0.029 & 12.4 \\
        70 & 100 & 1290 & 0.028 & 33.4 \\
        \bottomrule
    \end{tabular}
\end{table}

All three variants use every codebook entry, so the increase in OR is not caused by codebook collapse. The decreasing PPL instead indicates more concentrated token usage. Relative to the 26\%-OR variant, the higher-OR variants trade some reconstruction fidelity for label stability, while improving early and converged policy success. Neither OR nor PPL is a direct estimate of perturbation-induced entropy or multimodal mutual information.

\subsubsection{Impacts of Residual Grammar}

\begin{figure}
    \centering
    \includegraphics[width=0.95\linewidth]{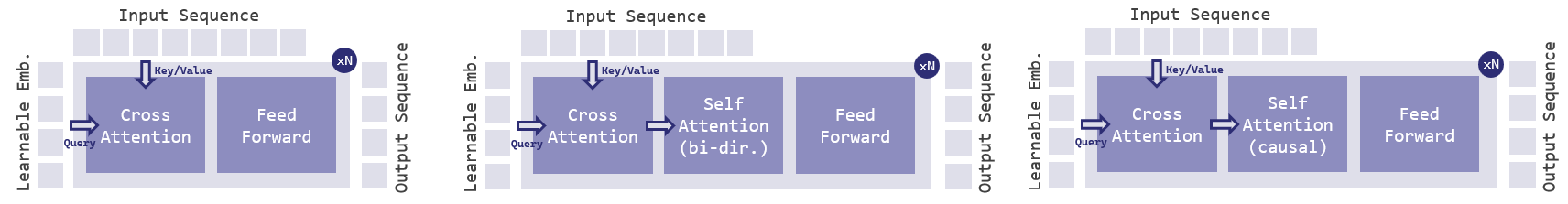}
    \caption{\textbf{Comparison of Perceiver architectures for Residual Grammar:} (left) Independent (cross-attention only), (middle) SA (with self-attention), and (right) Causal (with causal self-attention)}
    \label{fig:residual_grammar_arch}
\end{figure}

In \Cref{sec:vl_align_and_grammar}, we analyze the impact of \textit{residual grammar} on VLA training by comparing three tokenizer variants: \textbf{Independent}, \textbf{SA} (Self-Attention), and \textbf{Causal}. These variants are implemented by modifying the internal architecture of the Perceiver network. As shown in \Cref{fig:residual_grammar_arch}, the network structure using only cross-attention layers (\textbf{Independent}) ensures that each learnable embedding interacts exclusively with the input sequence. Consequently, the latent queries lack direct self-attention coupling, while their outputs may still be statistically dependent through the shared input action. By introducing an additional self-attention layer, the learnable embeddings interact with each other; specifically, we can employ either bidirectional attention (\textbf{SA}) or causal masking (\textbf{Causal}) to alter the dependency relationships between the output tokens.

To specifically analyze the impact of these variants on VLA training, we design a \textbf{perturbation experiment} in \Cref{fig:grammar}. The procedure is as follows: we sample vision-language-action tuples $(V, L, A)$ from the dataset and provide $V$ and $L$ as inputs to the trained VLA model to predict action tokens. During the autoregressive prediction process, we randomly shuffle the token at a specific position. The model then continues to predict the remaining tokens. Finally, the predicted action tokens are decoded into a predicted action $\hat{A}$, and we calculate the $L_1$ distance between the ground truth $A$ and the predicted $\hat{A}$. Our results show that the \textbf{Independent} variant maintains a stable and lower error compared to the \textbf{SA} and \textbf{Causal} architectures. This indicates that VLA performance using SA (full dependency) or Causal (temporal dependency) is inferior to the Independent variant. This is attributed to the fact that when predicting action tokens, the VLA model becomes overly reliant on preceding tokens, leading to reduced robustness.

\subsection{Details of Tokenizer Comparison Experiments}

\textbf{Action Tokenizers.} We evaluate the ActionCodec models introduced in \Cref{sec:actioncodec}, pre-trained on LIBERO \citep{liu2024libero}, BridgeData \citep{walke2023bridgedata}, and DROID \citep{khazatsky2024droid}. Training utilized a batch size of 8,192 and a learning rate of $2\times 10^{-4}$ for 100k steps. Following the FAST tokenizer \citep{pertsch2025fast}, our tokenizer encodes a fixed temporal window of 1 second. To accommodate heterogeneous control frequencies across datasets, we employ timestamp-based positional encodings to ensure temporal awareness. 

For baselines, we implement several mainstream schemes: (i) \textbf{Binning} \citep{goyal2025vla0}: Each action dimension is discretized into 1,000 bins. Due to its linear token growth relative to action dimensionality and horizon, a 1-second window (20Hz) would result in a prohibitive 140 tokens per chunk for 7-DoF LIBERO tasks; thus, we constrain its horizon to $T=8$. (ii) \textbf{String} \citep{hancock2025vlm2vla}: Actions are represented as raw Python list strings. Since this representation incurs an even higher token budget depending on precision and BPE encoding, we similarly adopt $T=8$ to maintain training tractability. (iii) \textbf{VQ-baselines}: We utilize official implementations\footnote{MiniVLA: \url{https://github.com/Stanford-ILIAD/openvla-mini}}\footnote{VQVLA: \url{https://github.com/xiaoxiao0406/VQ-VLA}} and checkpoints for MiniVLA \citep{belkhale2024minivla} and VQ-VLA \citep{wang25vqvla} after aligning data statistics. (iv) \textbf{FAST}: We utilize the officially released universal version \citep{pertsch2025fast}\footnote{FAST: \url{https://huggingface.co/physical-intelligence/fast}}.

\textbf{VLA Implementation.} Consistent with our validation experiments, all tokenizers are evaluated using a SmolVLM2-2.2B backbone without structural modifications. We perform full-parameter autoregressive fine-tuning; specific hyperparameters are summarized in \Cref{tab:vla_hyperparams_libero}.

\begin{table}
\centering
\caption{\textbf{VLA Training Hyperparameters.} Detailed configurations for the validation experiments on LIBERO-Goal.}
\label{tab:vla_hyperparams_validation}
\small
\begin{tabular}{l@{\hspace{8pt}}>{\raggedright\arraybackslash}p{0.60\linewidth}}
\toprule
\textbf{Hyperparameters} & \textbf{Values} \\
\midrule\midrule
GPUs & 8$\times$NVIDIA RTX 3090 (24 GB VRAM) \\
Learning Rate & $2\times10^{-4}$ peak (1k steps linear warmup, 30k steps cosine decay to $2\times10^{-5}$) \\
Global Batch Size & 128 (16 per GPU) \\
Training Steps & 30k (LIBERO-Goal only) \\
Input Modalities & 1$\times$ Third-person camera, 1$\times$ Wrist camera (concatenated) \\
Input Image Size & 224 $\times$ 448 \\
Use Proprioception & True \\
Image Augmentation & Random crop (87.5\%), Brightness/Contrast/Saturation (0.2), Hue (0.05) \\
\bottomrule
\end{tabular}
\end{table}

\begin{table}
\centering
\caption{\textbf{VLA Training Hyperparameters.} Detailed configurations for the tokenizer comparison experiments on LIBERO.}
\label{tab:vla_hyperparams_libero}
\small
\begin{tabular}{l@{\hspace{8pt}}>{\raggedright\arraybackslash}p{0.60\linewidth}}
\toprule
\textbf{Hyperparameters} & \textbf{Values} \\
\midrule\midrule
GPUs & 4$\times$NVIDIA A100 (80 GB VRAM) \\
Learning Rate & $10^{-4}$ peak (1k steps linear warmup, 30k steps cosine decay to $10^{-5}$) \\
Global Batch Size & 128 (32 per GPU) \\
Training Steps & 30k (applied to all four LIBERO task suites) \\
Input Modalities & 1$\times$ Third-person camera, 1$\times$ Wrist camera (concatenated) \\
Input Image Size & 224 $\times$ 448 \\
Use Proprioception & True \\
Image Augmentation & Random crop (87.5\%), Brightness/Contrast/Saturation (0.2), Hue (0.05) \\
\bottomrule
\end{tabular}
\end{table}

\begin{table}
\centering
\caption{\textbf{VLA Training Hyperparameters.} Detailed configurations for the Simpler-WidowX experiments.}
\label{tab:vla_hyperparams_simpler}
\small
\begin{tabular}{l@{\hspace{8pt}}>{\raggedright\arraybackslash}p{0.60\linewidth}}
\toprule
\textbf{Hyperparameters} & \textbf{Values} \\
\midrule\midrule
GPUs & 8$\times$NVIDIA H100 (94 GB VRAM) \\
Learning Rate & $2\times10^{-5}$ \\
Global Batch Size & 128 (16 per GPU) \\
Training Epochs & 4 \\
Input Modalities & 1$\times$ Third-person camera \\
Input Image Size & 224 $\times$ 224 \\
Use Proprioception & True \\
Image Augmentation & Random crop (90\%), Brightness/Contrast/Saturation (0.2), Hue (0.05) \\
\bottomrule
\end{tabular}
\end{table}

\begin{table}
\centering
\caption{\textbf{VLA Training Hyperparameters.} Detailed configurations for the real-world experiments on SO100 and xArm.}
\label{tab:vla_hyperparams_real}
\small
\begin{tabular}{l@{\hspace{8pt}}>{\raggedright\arraybackslash}p{0.60\linewidth}}
\toprule
\textbf{Hyperparameters} & \textbf{Values} \\
\midrule\midrule
GPUs & 4$\times$NVIDIA A100 (80 GB VRAM) \\
Learning Rate & $5\times10^{-5}$ peak (1k steps linear warmup, 30k steps cosine decay to $5\times10^{-6}$) \\
Global Batch Size & 64 (16 per GPU) \\
Training Steps & 30k \\
Input Modalities & \textbf{SO100:} 1$\times$ Third-person camera, 1$\times$ Wrist camera (concatenated), \textbf{xArm:} 1$\times$ Third-person camera \\
Input Image Size & 224$\times$448 for SO100, 224$\times$224 for xArm \\
Use Proprioception & True \\
Image Augmentation & Brightness/Contrast/Saturation (0.2), Hue (0.05) \\
\bottomrule
\end{tabular}
\end{table}

\textbf{Datasets and Benchmarking.} A single VLA model is trained across all LIBERO task suites simultaneously. We observe empirical convergence for all variants by 30k steps, beyond which performance gains become negligible. Following the official LIBERO evaluation protocol, we report results based on 50 trials per task, initialized from the 50 predefined environment states.

\begin{figure}
    \centering
    \includegraphics[width=0.95\linewidth]{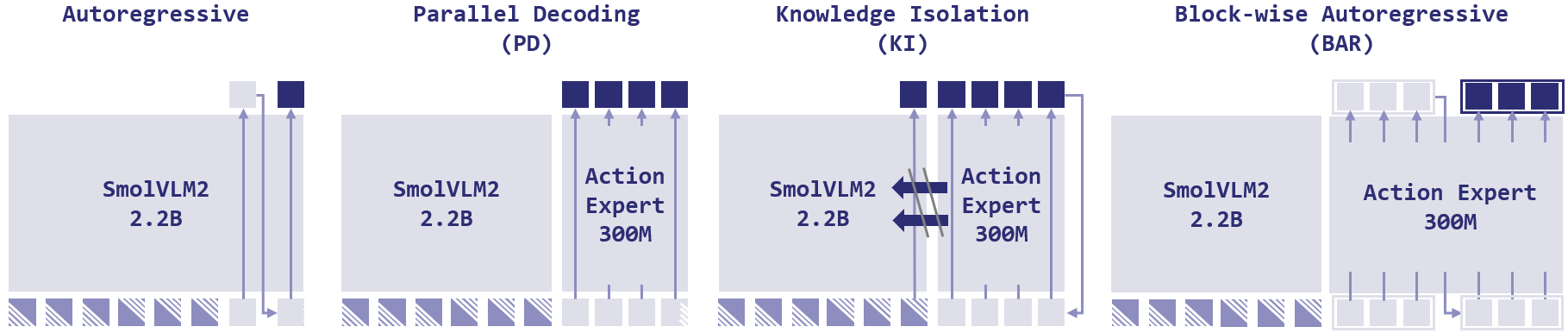}
    \caption{\textbf{VLA paradigm architectures.}}
    \label{fig:paradigms}
\end{figure}

\begin{figure}[t]
\centering
\begin{tcolorbox}[title=Prompt for VLM, fonttitle=\bfseries, colback=gray!5, colframe=gray!50, width=0.95\linewidth]
\begin{small}
\begin{verbatim}
[
  {
    "role": "system",
    "content": "Analyze the image and predict robot actions."
  },
  {
    "role": "user",
    "content": [
      {"type": "image"},
      {"type": "text",
       "text": "**State**: {state}, **Task**: {task}."}
    ]
  },
  {
    "role": "assistant",
    "content": [
      {"type": "text",
       "text": "{'action': <|action_n|> ... <|action_m|>}"}
    ]
  }
]
\end{verbatim}
\end{small}
\end{tcolorbox}
\caption{\textbf{Prompt template for VLM fine-tuning.}}
\label{fig:vlm_prompt}
\end{figure}

\subsection{Details of Paradigm Adaptation Experiments}
\label{appendix:paradigm}

In \Cref{sec:paradigms_exp}, we explore the compatibility of \textbf{ActionCodec} with three prevailing VLA paradigms. Since each paradigm requires a specialized attention mask design for action tokens, it often prevents the VLM backbone from utilizing \textit{Flash Attention} \citep{dao2024flashattention}, which significantly degrades training efficiency. To circumvent this, following the architecture of $\pi_0$ \citep{black2024pi0}, we introduce a 300M-parameter \textbf{action expert} to handle action token prediction. This expert receives the KV cache from the VLM backbone and employs independent feed-forward layers. We provide the architecture in \Cref{fig:paradigms} and the implementation details are as follows:

\textbf{Parallel Decoding (PD).} The action expert utilizes bidirectional attention and takes $n$ learnable \texttt{[BOS]} tokens as input to directly predict the logits of $n$ action tokens in a single forward pass. During inference, we perform a simple \textit{argmax} on the logits to obtain tokens. We observed that \textit{argmax} yields performance comparable to probabilistic sampling, likely because ActionCodec produces highly condensed semantic information with naturally low perplexity. Furthermore, PD achieves success rates similar to naive autoregression. This can be attributed to the token independence of ActionCodec, which makes modeling inter-token correlations via autoregression redundant. Compared to naive autoregression (requiring $n$ forward passes), PD significantly reduces inference latency by requiring only one.

\textbf{Knowledge Isolation (KI).} The action expert is implemented as a bidirectional attention network within a \textit{flow-matching} framework. It takes noisy actions, denoising timestamps, and the VLM's KV cache as input to predict the velocity field. During training, both the VLM's autoregressive loss and the expert's flow-matching loss are optimized simultaneously, though the gradient flow from the expert to the VLM is severed. While the performance of this KI variant is slightly lower than the naive autoregressive baseline, it remains superior to KI training using the FAST tokenizer. This suggests that while ActionCodec is highly effective, the KI paradigm may be better suited for large-scale VLA pre-training, where the VLM can be enriched with broad robotic priors, rather than fine-tuning on specific benchmarks.

\textbf{Block-wise Autoregressive (BAR).} The action expert utilizes a block-wise causal mask, where tokens within the same block are mutually visible, while blocks can only attend to preceding blocks. Each block corresponds to tokens from a single level of the RVQ codebooks. Due to the residual nature of RVQ, BAR with a single forward pass is equivalent to PD. Additional passes iteratively refine prediction accuracy; even if initial predictions are noisy, the error-damping property of the residual structure ensures stability. In our experiments, the BAR variant achieved the highest performance, establishing a new \textbf{SOTA} on the LIBERO benchmark for models without robotics-specific pre-training.

\begin{figure}
    \centering
    \includegraphics[width=0.3\linewidth]{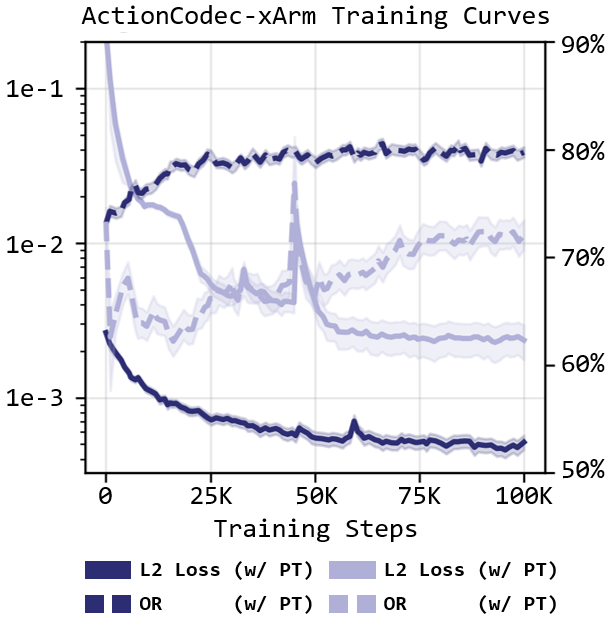}
    \caption{\textbf{Learning curves of ActionCodec with and without pre-training.} Compared to random initialization, initializing with pre-trained ActionCodec weights and fine-tuning on novel robotic data results in significantly lower L2 loss (higher reconstruction accuracy) and a higher Overlap Rate (OR). These results demonstrate that the pre-trained ActionCodec possesses a unified and semantically rich latent space, enabling rapid adaptation to new robotic action patterns.}
    \label{fig:xarm_curves}
\end{figure}

\section{Details of Soft-Prompt}\label{appendix:soft_prompt}

\begin{figure}
    \centering
    \includegraphics[width=0.7\linewidth]{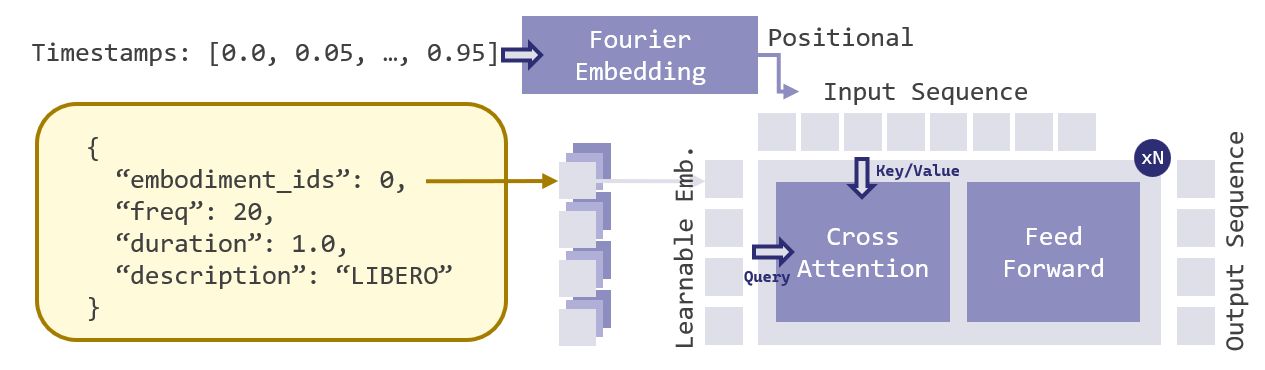}
    \caption{\textbf{Illustration of the Soft-Prompt integration within the Perceiver architecture.} Learnable embeddings represent embodiment IDs, while fourier embeddings provide temporal grounding for varying control frequencies.}
    \label{fig:soft_prompt}
\end{figure}

In \Cref{sec:actioncodec}, we employ a \textbf{soft-prompt} mechanism to facilitate multi-embodiment data integration and cross-platform knowledge transfer. This approach is highly compatible with the \textbf{Perceiver} architecture, as the Perceiver fundamentally requires learnable embeddings to serve as the input for the query side of the Transformer-based network. Specifically, as shown in \Cref{fig:soft_prompt}, we have configured information for all embodiments included in the pre-training dataset, documenting their respective control frequencies, the action durations encoded by the tokenizer, and their unique indices. We assign a dedicated learnable embedding to each embodiment index. For the input components involving action timestamps, specifically the key/value inputs of the encoder and the query inputs of the decoder, we calculate the timestamp for each token based on the control frequency and action duration. These timestamps are then processed via fourier embedding to serve as positional encodings. This design ensures that our tokenizer can not only distinguish between different embodiments but also maintain a consistent temporal perception of motion across various platforms.

\section{Details of RVQ Post-training}\label{appendix:rvq_ft}

\begin{figure}
    \centering
    \includegraphics[width=0.95\linewidth]{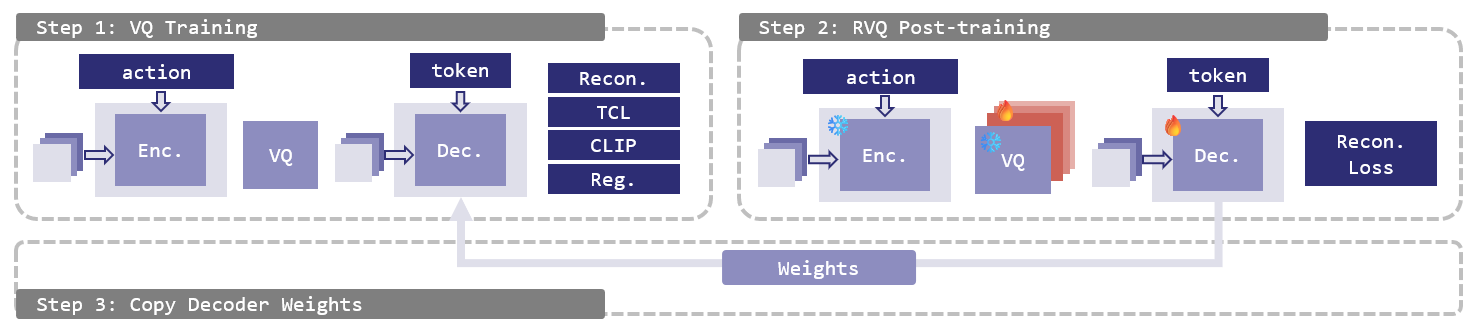}
    \caption{\textbf{Procedure of RVQ post-training.}}
    \label{fig:rvq_ft_procedure}
\end{figure}

As discussed in \Cref{sec:actioncodec}, we find that RVQ post-training is an effective method for reducing reconstruction error at no additional cost while maintaining a high Overlap Rate (OR) and vision-language alignment. In our experimental setup, as shown in \Cref{fig:rvq_ft_procedure}, we first train a VQ-based ActionCodec with $n=16$ tokens. We then initialize an RVQ variant with $n=16$ and a residual depth of $L=3$, inheriting the encoder and primary codebook weights from the VQ model and freezing them. Post-training is performed using only reconstruction loss and RVQ objectives. Since the encoder and the first codebook are frozen, the output tokens from the first codebook of the RVQ model remain identical to those produced by the original VQ version. By simply copying the post-trained RVQ decoder weights back to the VQ model, we enhance reconstruction fidelity without altering the action tokenization procedure. In our evaluations, ActionCodec-BAR utilizes the full RVQ tokenizer, whereas all other models employ the VQ-version tokenizer enhanced by the post-trained RVQ decoder.

\section{Scope of the Information-Theoretic Analysis}\label{appendix:artifact_entropy}

\subsection{Deterministic and Perturbation-Induced Entropy}
The cross-entropy decomposition and the entropy/mutual-information identities in \Cref{sec:before_method} are exact. However, these identities alone do not guarantee that optimizing individual terms improves the learned policy. Our design recommendations are empirical guidance supported by the controlled experiments, and the relevant properties may change together.

Let $\mathcal{T}$ denote the deterministic action encoder followed by discrete quantization, so that $C=\mathcal{T}(A)$. For this mapping, $H(C|A)=0$ and $I(C;A)=H(C)$. To discuss sensitivity to sensor noise or control jitter, define a separate random token sequence $C_{\epsilon}=\mathcal{T}(A+\epsilon)$, with $\epsilon\sim\mathcal{N}(0,\sigma^2 I)$, and its perturbation-induced entropy
\begin{equation}\label{eq:perturbation_entropy}
    H_{\epsilon}(C_{\epsilon}|A)
    =\mathbb{E}_{a\sim P_A}\!\left[H_{\epsilon}\bigl(\mathcal{T}(a+\epsilon)\bigr)\right].
\end{equation}
The inner entropy is taken over the perturbation distribution at a fixed action $a$. This quantity describes sensitivity to quantization boundaries; it does not replace $H(C|A)$ in the deterministic entropy identity. Interpreting temporal neighbors as a local-stability probe assumes that their underlying motions and contexts are similar. OR (\Cref{eq:overlap_rate}) measures the observable consistency of their tokens under this assumption; it does not numerically estimate entropy. The action-perturbation InfoNCE experiment in \Cref{appendix:or_controls} separately probes local stability by pairing $A$ with $A+\eta$.

\subsection{Capacity Subject to Fidelity}\label{appendix:capacity_scope}
The exact capacity bound $I(C;A)\le H(C)\le\sum_k H(c_k)\le n\log_2 S$ specifies a representational upper bound, not a policy-performance optimum. A practical bottleneck objective is
\begin{equation}
    \min_{\mathcal{T},\mathcal{G}} H(C)
    \quad\text{subject to}\quad
    \mathbb{E}[d(A,\hat A)]\le\varepsilon,
    \qquad \hat A=\mathcal{G}(C),
\end{equation}
where $d$ is a reconstruction distortion and $\varepsilon$ is an acceptable tolerance. Reducing capacity too far violates the fidelity constraint. Our $n=16,S=2048$ configuration is an empirical trade-off, rather than a unique optimum; $S=2048$ also matches the official universal FAST vocabulary. Holding OR at 40\% in the $n,S$ sweeps reduces one confound, but reconstruction quality and token-usage distributions can still vary.

\subsection{Alignment and Architectural Coupling}\label{appendix:alignment_scope}
The chain rule in \Cref{sec:vl_align_and_grammar} decomposes complementary sources of predictive information: multimodal context and the autoregressive prefix. We do not directly estimate either mutual information, which would require the unknown joint distribution of tokens, observations, language, and history. Attention visualizations and CLIP/TCL comparisons serve as qualitative context-alignment proxies; the shuffled-prefix experiment measures token-error robustness, rather than mutual information.

TCL provides a practical, indirect route toward vision/action alignment through temporal correspondence, without the direct vision/action CLIP computation. It is therefore cheaper than directly aligning action features with visual features, but it may also capture demonstration-specific patterns. In \Cref{fig:vl_align}, CLIP emphasizes instruction-relevant objects whereas TCL emphasizes dataset-specific demonstration patterns. These observations support using the objectives as practical training tools, without establishing a numerical increase in mutual information.

Finally, \textit{independent tokens} refers to the cross-attention-only architecture. All latent queries attend to the same input action, so statistical independence is neither imposed nor claimed. The intended balance is to retain shared action structure while limiting reliance on prior output tokens. The prefix-error results support this choice for the tested architectures and do not establish a universal requirement for all action tokenizers.

\section{Additional Experimental Results and Comparison Protocols}

\subsection{Large-Scale VLA Pre-training and Transfer}\label{appendix:large_scale}
In addition to the SmolVLM2 experiments without VLA pre-training, we pre-train ActionCodec+Qwen3.5-2B on 15K hours of robotic data spanning 14 embodiments. We use a $\pi_{0.5}$-style knowledge-isolation (KI) setup with ActionCodec supervision. \Cref{tab:large_scale_transfer} reports benchmark-average success for downstream task fine-tuning and zero-shot transfer. These results concern a VLA-pre-trained model and are separate from the models without VLA pre-training in \Cref{tab:libero_results_refined_final}.

\begin{table}[htbp]
    \centering
    \caption{\textbf{Large-scale VLA pre-training and transfer.} Benchmark-average success rates (\%). FT denotes downstream task fine-tuning; ZS denotes zero-shot evaluation. A dash denotes an unreported result.}
    \label{tab:large_scale_transfer}
    \small
    \begin{tabular}{lccc}
        \toprule
        Method & Simpler-WidowX FT & LIBERO FT & DROID ZS \\
        \midrule
        $\pi_0$-FAST & 32.1 & 85.5 & -- \\
        $\pi_{0.5}$ & 57.1 & 96.8 & 57.5 \\
        MolmoAct2-DROID & -- & -- & 52.0 \\
        ActionCodec (Qwen3.5-2B) & \textbf{87.7} & \textbf{98.2} & \textbf{82.5} \\
        \bottomrule
    \end{tabular}
\end{table}

ActionCodec obtains the highest reported average in all three columns. This extends the evidence to both downstream fine-tuning and zero-shot transfer at larger data scale. These benchmark averages support transfer on the evaluated settings; they do not establish broad in-the-wild generalization.

\subsection{Baseline Fairness and Encoding Cost}\label{appendix:fairness_efficiency}
The experiments in \Cref{sec:before_method} control tokenizer properties within the VQ family, while \Cref{tab:libero_results_refined_final,tab:tokenizer_efficiency} compare complete methods in official or practical configurations. Enforcing equal token counts would change FAST and the other baseline methods. Token budget, action horizon, OR, policy latency, and action throughput are therefore reported explicitly in \Cref{tab:tokenizer_efficiency}. FAST and ActionCodec both cover 20 action steps (one second) on LIBERO, with average token budgets of 24.3 and 16, respectively. Other baselines use their reported horizons; these comparisons should not be interpreted as matched-budget or matched-horizon ablations.

We use the officially released universal FAST tokenizer in both the validation and full comparison experiments. Soft decoding only pads or truncates BPE-decoded sequences of invalid length, preserving valid outputs. This repair improves FAST's early-training usability without altering its tokenizer. The backbone-size comparison also favors FAST: ActionCodec with 256M/500M backbones reaches 92.3/92.5\% on LIBERO, exceeding FAST with a 2.2B backbone (90.6\%). The tokenizer gain therefore cannot be explained by a larger ActionCodec backbone.

\begin{table}[htbp]
    \centering
    \caption{\textbf{Action encoding cost.} Per-chunk encoding time at batch size 1024. These measurements concern tokenizer encoding, separately from policy generation latency in \Cref{tab:tokenizer_efficiency}.}
    \label{tab:encoding_cost}
    \small
    \begin{tabular}{lc}
        \toprule
        Tokenizer & Encoding time per chunk (ms) \\
        \midrule
        FAST & 0.11 \\
        ActionCodec & 0.09 \\
        \bottomrule
    \end{tabular}
\end{table}
The small measured encoding cost supports using training steps as a practical measure of convergence under the shared VLA training configuration. It does not establish identical total wall-clock training cost across token budgets.

\subsection{Optimization and Overfitting Evaluation}\label{appendix:optimization_metrics}
Raw token NLL is not directly comparable across tokenizers with different vocabularies and sequence lengths. We instead track task success throughout training (\Cref{fig:teaser,fig:toy_example_ablation}) to assess convergence and late-stage degradation. Closed-loop rollout success tests deployment behavior: predicted actions change subsequent observations, allowing small errors to accumulate under distribution shift. This complements teacher-forced validation loss. Late-stage rollout degradation in \Cref{fig:toy_example_ablation} is the empirical basis for our discussion of overfitting; latent-space visualizations are supporting qualitative evidence.


\end{document}